\documentclass{article}

\usepackage{mathptmx}       
\usepackage{helvet}         
\usepackage{courier}        
\usepackage{type1cm}        
\usepackage{makeidx}         
\usepackage{graphicx}        
\usepackage{multicol}        
\usepackage[bottom]{footmisc}

\usepackage[english]{babel}
\usepackage{fontenc}

\usepackage[round]{natbib} 
\usepackage{url}

\usepackage{bm} 
\providecommand{\shortcite}[1]{\citeyearpar{#1}}
\renewcommand{\cite}[1]{\citep{#1}}

\setcitestyle{square}

\newcommand{\og}[1]{``}
\newcommand{\fg}[1]{''}

\makeindex             
                       
\newcommand{\scr}[1]{\scr #1}
\def\ss{{\cal S}}
\def\as{{\cal A}}
\def\hs{{\cal H}}
\def\nit{\hbox{\it I\hskip -2pt N}}
\newtheorem{theorem}{Theorem}

\newcommand{\pierre}[1]{\footnote{ \textbf{PM:} #1}}
\renewcommand{\pierre}[1]{}
\usepackage{a4wide}

\input{scrload.tex}

\errorstopmode
\begin{document}
                       
\title{Reasoning about Action and Change\thanks{This is an accepted manuscript
    of the chapter Reasoning about Action and Change of the book A Guided Tour of Artificial Intelligence Research, 1 / 3, Springer International Publishing, pp.487-518, 2020}}\label{volume-1-chapitre-12}
\author{Florence Dupin de Saint-Cyr$^1$, Andreas Herzig$^1$, J\'er\^ome
  Lang$^2$, Pierre Marquis$^3$\\
\small $^1$ IRIT-CNRS. Universit\'e Paul
  Sabatier, Toulouse, France\\
\small $^2$ CNRS, Universit\'e Paris-Dauphine, PSL Research University, LAMSADE, Paris, France\\
\small$^3$ CRIL-CNRS, Universit\'e d'Artois \& Institut Universitaire de France, Lens, France}

\maketitle


\abstract{This chapter presents the state of research concerning the formalisation of an agent reasoning
about a dynamic system which can be partially observed and acted upon.
We first define the basic concepts of the area: system states, ontic and epistemic actions,
observations; then the basic reasoning processes:  prediction, progression,
regression, postdiction, filtering, abduction, and extrapolation. We
then recall the classical action representation problems and 
show how these problems are solved in some standard frameworks.
For space reasons, we focus on these major settings: 
the situation calculus, STRIPS and some propositional action languages, 
dynamic logic, and dynamic Bayesian networks.
We finally address a special case of progression, namely belief update.}

\section{Introduction}


In this chapter, we are interested in \emph{formalizing the reasoning of a single agent} \index{agent}  
who can make \emph{observations} \index{observation} on a \emph{dynamic system}
\index{system!dynamic} and considers \emph{actions} to perform on it.
Reasoning about action \index{action} and change \index{change} is among the
first issues addressed within Artificial Intelligence (AI); especially, it was
the subject of the seminal article by McCarthy and Hayes \shortcite{McHa69}.
Research in this area has been very productive until the late 1990s. 
Among other things, 
solutions to the various problems to be faced when dealing with action representation were put forward and
a classification of action languages according to their expressive power was undertaken. 
Moreover,  much progress towards the automatization of reasoning about action and
change was made, for example through the design and the evaluation of algorithms implementing the reasoning processes of the main action languages
and the investigation of the computational complexity of such processes.


The reasons why an agent may wish to act in order to modify the current state of a dynamic system or to learn more about it are numerous. For example,
the goal can be to change the system into a configuration that the agent prefers
over the actual one (such as moving a robot from a location to another one), or even into an optimal configuration.
Alternatively, the objective can be to ensure that a certain property of the dynamic system is maintained, or 
that its successive states do not deviate too much from a normal path.
The latter is for example the case when one wants to supervise and control a physical system, 
such as a furnace or that of a patient in an intensive care unit.
Such scenarios involve concepts (state, action, observation,
etc.) and processes connecting them (planning, prediction, explanation, etc.).


By `formalizing', we first mean \emph{modeling the concepts and
processes} that are considered in such scenarios (the purpose is to define
them rigorously from a mathematical point of view) and then 
\emph{representing} them (that is, specifying how the information are
encoded) and \emph{automating} (designing algorithms suited to the processes under consideration).
Note that there are typically two main reasons for modeling a dynamic system: {\em controlling} 
it (see Chapter 10 of Volume 2 about planning), 
and obtaining more information about it, for
diagnosing it \index{diagnosis} or supervising it \index{supervision} (see Chapter 31 of this volume).
Once modeled, the same concept can be associated with several
representations. If the choice of a specific model typically depends on the available pieces of 
information and what one wants to do with them, the choice of a representation (suited to a given model) 
is based on other criteria, such as computational efficiency and succinctness.

\section{Reasoning about Action: Models}

\subsection{Basic Concepts and the Corresponding Models}
%

%
%

In this section, we define some mathematical notions corresponding to
the key concepts considered in reasoning about action and change.

The {\em model} of a reasoning process on  \emph{a dynamic system} 
\index{system!dynamic} can be divided in two parts: the  model of the system (with its own dynamics) 
and the  model of the agent (including her knowledge about the system).
Sandewall \shortcite{Sandewall95} has developed a taxonomy of reasoning problems
on dynamic systems, and the remainder of this chapter elaborates on it.

Throughout the chapter, we assume that {\em time} \index{time} is 
discrete 
(which is a common assumption in Artificial Intelligence). The {\em horizon}  \index{horizon}
of the process is the set $\hs$ of relevant steps for
controlling and observing it. It can be finite $(\hs = \{0, \ldots, N\}$ with
$N\in \nit$) or infinite
($\hs = \nit$); a degenerate case of a finite horizon is when there is only one change step
($\hs = \{0,1\}$).

A {\em state} \index{state} is the description of the system at a given time point. Unless stated
otherwise, \emph{the set of all states}, denoted by $\ss$, will be assumed
finite. A {\em state trajectory} \index{trajectory} is a sequence of elements of $\ss$, indexed by elements of $\hs$.
The system states at the different time points of $\hs$ may only partially known by the agent.

The specification of a reasoning process \index{reasoning!temporal} on a dynamic system 
requires first the \emph{beliefs} \index{belief}  of the agent about the state of the system at different time points
(including the initial time point) and about the general laws that govern the evolution of the system. 
Thus, we first have to choose a model for uncertain belief states\index{uncertainty}\index{state!belief}.
For space reasons, we will focus only on two uncertainty models in the following: 
the {\em binary model}, where belief states $b$ are \emph{non-empty subsets of $\ss$} and
the {\em Bayesian model}, where belief states $b$ are \emph{probability distributions} \index{probability} 
on $\ss$.\footnote{There are many other uncertainty models 
that should be mentioned but will not be, for
space reasons - they include ordinal models, where belief states
and action effects are modeled as pre-orders
\index{preference} over $\ss$, possibilistic models \index{logic!possibilistic}
that are similar in spirit to them, non-Bayesian probabilistic models, where a belief state is a family of 
probability \index{probability} distributions, etc. (see Chapter 3 of this volume).}

Transitions \index{transition} from one state to another are triggered by 
{\em events}\index{event}. These events generally change not only the state of the system, but also
the beliefs of the agent.
An {\em action} \index{action} is a special event triggered by an agent.
The agent has a model of each \emph{action} available to her. 
The set of actions available to the agent is denoted by $\as$, and 
is supposed to be finite.
The agent can also have a model for \emph{exogenous events}, 
 which are phenomena whose dynamics are similar to actions
 but which are not triggered by the agent. They are triggered
by nature or possibly by other agents more or less well-identified 
(i.e., whose identity may be imperfectly known), 
and their occurrences are {\em a priori} not known  by
the agent. We distinguish between the {\em action type}
$\alpha$ (defined very generally) from the action occurrence(s) at one or
more time point(s): a given action may have no occurrence in
an instance of a problem, or may have one or several occurrences.
Actions have two types of effects: {\em ontic (physical) effects}, on the world, 
and {\em epistemic effects}, on the beliefs of the agent. 
Epistemic effects can either be caused by her projection of the physical effects of the performed action 
(for instance, if I know that the action ``delete file F" has the effect that file F no longer appears 
on my computer, then, when I execute this action, the resulting belief state
is such that I know that F no longer is on my computer) or from {\em observations} or any form of feedback 
(for instance, if after trying to turn the light on by flipping the switch, I observe that the light is off, then,
in my new state of belief, I know that the bulb is broken or that the power is off).


Actions have generally two types of effects at once (as in the case of the ``switch'' action above).
Some actions, referred to as {\em purely epistemic actions}\index{action!epistemic}, have only epistemic effects, and no effect on the state of the world; 
for example, measuring a temperature, or querying a database.
Other actions, referred to as {\em purely ontic actions} \index{action!ontic} have epistemic effects (it is hard to imagine actions without any epistemic effect, apart from the action ``do nothing"), but these epistemic effects 
are the simple projection, by the agent, of what she knows about the ontic effects of the
action (as for the action ``delete file F" above). In
other words, a purely ontic action gives {\em no  feedback} to the 
agent: her belief state after the execution of such an action coincides
with the belief state she could foresee before executing the action
(``what you foresee is what you get").
Every action
can be decomposed in a unique way into a purely ontic action
and a purely epistemic action; without loss of generality, we can thus assume that each
available action is either purely ontic or purely epistemic (and we will make such an assumption in the rest of the chapter, unless  stated otherwise).

%

Let us start by describing \emph{purely ontic actions}.
The effects of a purely ontic action $\alpha$ are defined by
a {\em transition system} \index{transition} between the states of the world, modeled as a
{\em binary relation} ${\bm R_\alpha}$ over $\ss$.

The simplest case is when actions are {\em deterministic and always executable}\index{action!deterministic}\index{action!executable}. 
In this case, the transition system
of $\alpha$ is a mapping $R_\alpha$ from $\ss$ to $\ss$. 
An action has {\em  conditional effects} \index{action!conditional effect} 
if the resulting state after its execution depends on the state before
its execution. For example, the action ``switch off the light" 
may be regarded as deterministic and unconditional (if we
assume that it always has the effect that the lamp is off
after its execution). ``Toggle the switch'' can be considered as deterministic, and with
conditional effects since its effects depend on the state (``on'' or ``off'') of the light before
the execution of the action.

More generally, actions are not always
executable: there can be states $s$ such that
$R_\alpha (s) = \emptyset$; actions can also be {\em non-deterministic}:
there are states $s$ such that $R_\alpha (s)$ contains more than one
element. For example, the action ``delete file F" is not executable 
if the file does not exist; in this case, the modeler will define
the effect of the action only for states where the file exists, and executing the action will be forbidden
in the other cases. Another model would make advantage of an action with conditional effects,
where the transition associated to the action would be associated with the identity
relation in situations where file F does not exist, and would lead to states where the file is deleted otherwise.

In the non-deterministic case, the transition model chosen depends on the nature of
the uncertainty one wants to deal with; with each initial state is associated 
a belief state on the subsequent states. Note that choosing a deterministic  
or a non-deterministic model for a system may depend on the knowledge and the goals of the modeler:
the action ``turn the computer off" can be considered as
non-deterministic for an agent who is not a computer scientist (since it may happen that after
the execution of the action the computer is still on) but as deterministic, with conditional effects,
for an expert in computer science (since this expert will be able to determine
the cases where the computer stays on after being turned off).
Modeling a dynamic system as a transition system \index{transition} between states  amounts to making
the implicit assumption that the system is Markovian.\footnote{A system is Markovian if the transition of 
the system to any given state depends only on the current
state and not on the previous ones.}
Such an assumption can be made without loss of generality 
by considering more complex states
(encoding state trajectories). For the sake of brevity, we will stick to the following 
two models: the binary non-deterministic model and the stochastic model.

%

In the binary non-deterministic model, the transition system
of an action $\alpha $ is a mapping $R_\alpha$ from $\ss$ to $2^\ss$
(or to $2^\ss \setminus \{\emptyset\}$, when $\alpha$ is always
executable). For example, if the system states are
$\ss = \{c\_on, c\_stand\_by, c\_\mbox{\textit{off}}\} $ (representing the activity of a computer: ``on'',
``stand-by'' or ``off'') then the action of ``shutting down the computer'' may be modeled as
$R_ {shut\_down} (c\_on) = \{c\_on, c\_\mbox{\textit{off}}\}$,
$R_ {shut\_down} (c\_stand\_by) = R_ {shut\_down} (c\_\mbox{\textit{off}}) = \emptyset$ (meaning that one
can ``turn off the computer'' only if it is ``on'', and in this case, it is not sure 
that the ``shut down'' action succeeds). Note that if $\alpha $ is a purely epistemic action, then
$R_\alpha (s) = \{s\} $ for all $s$.

In the stochastic model (here, the Bayesian model for uncertainty), 
$R_\alpha$ is a stochastic matrix, i.e., a family of 
probability distributions $p(. | s, \alpha) $ for $s \in \ss$, 
where $p (s' | s, \alpha) $ is the probability \index{probability} to obtain 
the state $s'$ after the execution of $\alpha$ in $s$. In this model, it is possible
to specify how much the ``shut down'' action succeeds;  thus, $R_{shut\_down}$ 
could be represented by $p(c\_on$ $|$ $c\_on$, $shut\_down)$ $=$ $ 0.1$,
$p(c\_\mbox{\textit{off}}$ $|$ $c\_on$, $shut\_down)$ $=$ $0.9$,
$p(c\_stand\_by$ $|$ $c\_on$, $shut\_down)$ $=$ $0$.


\emph{Epistemic effects} of actions \index{action!epistemic} are expressed in terms of feedback. 
The actions that the agent decides to execute do not depend directly on the system state (which may be unknown to the agent)
but  on the agent's beliefs (and in particular, on what has been observed earlier).
Ideally, the current state and what the agent may observe coincide.
In this case, the belief state of the agent is perfect, but this hypothesis
reflects an ideal case and does not often hold. In order to define the 
epistemic effects of actions, an {\em observation space} \index{observation} ${\bm \Omega}$ can be introduced in the model.
This space, unless otherwise indicated, is supposed finite. 
The observations are the feedback given by the system
and each observation at a given time point from the horizon is
a projection of a state (not necessarily totally
observed) of the system. The observations are called \emph{reliable} if this 
projection corresponds to the \emph{actual state} of the system (unreliable observations
can arise from faulty sensors, for example).

%

Taking observations into account concerns two distinct stages:
the off-line stage when the decision policy is generated, and
the on-line stage when it is exploited (i.e., when the plan \index{plan} is executed).
During the off-line stage, the agent who generates the policy takes advantage of 
her knowledge about the observations which could be made at the on-line stage.
During the on-line phase, the actions which are triggered by the agent typically depend on the
observations which are effectively made. Note that nothing prevents from having
two distinct agents (one who computes a decision policy and another one who
executes it).

Two assumptions corresponding to two extreme cases are commonly made: when
the system is {\em fully observable}, the observation space is
identical to the state space: when she generates a decision policy, the agent knows 
that at the on-line stage, the actual state of the system will be known exactly at each
time point; when the system is {\em non-observable}, the observation space is a singleton
$\{o^*\}$, where $o^*$ is a fictitious observation  (the empty observation): the system
gives no feedback.


When none of these two extreme assumptions hold, one faces the
more general situation of {\em partial observability}, where observations
and system states are constrained by an observation-state {\em correlation structure}, 
the definition of which varies with the uncertainty model under consideration. 
In general, each state  $s$ and each \emph{epistemic action} $\alpha$ correspond to
a {\em belief state} ${\bm O}_{\bm \alpha} ({\bm s}) $ over the space of
observations, representing the {\it prior} beliefs about the observations obtained
when action $\alpha$ is performed in state $s$. In the binary model for uncertainty, 
a family of sets ${\bm O}_\alpha (s) $ for $s \in \ss$ is thus considered, where
$O_\alpha (s) $ {\em is a non-empty set of $\Omega$}; so $o \in O_\alpha (s) $
reflects the fact that $s$ is a state compatible with the observation $o$ which results from the execution of 
$\alpha$. If $\alpha$ is purely ontic, then $O_\alpha (s) = \{o^*\} $ for all $s$.
In the Bayesian model for uncertainty\index{uncertainty}, 
the {\it feedback} is modeled by a probability distribution $p (. | s, \alpha)$ on $\Omega$ where
${\bm p (o | s, \alpha)}$ {\em is the probability} \index{probability} \emph {to observe $o$ when $\alpha$ 
is executed in $s$}.


In this section, we assumed that only one action at a time 
can be executed. In some problems, it is natural to
perform several actions in a {\em concurrent} way\index{action!concurrent}.
This requires to be able to define the effects of combinations of actions;
for reaching this goal, the same models as previously considered can be used,
viewing every possible combination of actions as a specific action.
A typical example \cite{Thielscher95}  is the one of a 
table that can be lifted by the right side or by the left side: the two actions performed in sequence and independently %
do not have the same effect as when they are executed simultaneously, especially when a glass of water is on the table!




\subsection{Types of Reasoning and their Implementations}

Reasoning on a dynamic system requires to take account of a time horizon, the \emph{prior} beliefs on the system (general laws of the domain and action effects),
the occurrences of actions at some time points, and the
observations at given time points (it is a simplified model -- see
\cite{Sandewall95} for a more general one, where, in particular, the
actions can have a duration). We are now going to approach some specific types of reasoning
implying reasoning on a dynamic system\index{reasoning!temporal reasoning},
as well as their implementation by means of algorithms.

\subsubsection{Prediction and Postdiction with Ontic Actions}
\index{reasoning!temporal reasoning}
{\em Prediction} \index{prediction} (also called \emph{projection}) consists in determining, according to one
initial state of belief $b$ and the description of a purely ontic action $\alpha$,
the new state of belief $b'$ resulting from the application of $\alpha$ in $b$. 
The transformation of a state of belief into another one by an action is called
{\em progression} \index{progression}; noted $b'= prog(b, \alpha)$. 
Of course, the formal definition of $prog$ depends on
the nature of the space of the beliefs (static and dynamic), thus it depends on
the chosen representation of uncertainty.
In the simplest case (that of classical planning) where
belief states are perfect, actions are deterministic  and always
achievable, each $prog (. , \alpha) $ is an application mapping a state
to another one. In the binary nondeterministic model, a state $s'$ is possible after the execution
of $\alpha$ in the state of belief $b \subseteq \ss$ if there exists a possible state
$s$ in the whole set of states corresponding to the initial belief
$b$, such as $s'$ is a possible result
of $\alpha$ in $s$, i.e. $prog (b, \alpha) = \bigcup_{s \in b} R_\alpha(s)$.
In the probabilistic model, the obvious choice is obtained by identifying the
model of the process to a Markov chain
\index{Markov decision process (MDP)} : $prog(b, \alpha)$ is the
probability distribution \index{probability}
$b'$ on $\ss$ defined by $b'(s') = \sum_{s \in \ss}b(s)p (s'|s, \alpha) $ (where
$p (. | s, \alpha) $ is the probability distribution associated with $R_\alpha$).


The second type of reasoning is {\em postdiction}\index{postdiction}.
It consists in determining, according to one
final state of belief $b'$ and the description of a purely ontic action
$\alpha$ which has just been carried out, the state of belief $b$ before
the action was done. This transformation of a state of belief into
another, is sometimes also called {\em regression}
\index{regression} or
{\em weak regression}; noted $b = reg_w (b', \alpha) $. 
The weak regression corresponds to the progression by the reverse action
of $\alpha$ (noted $\alpha^{- 1}$), which transition system $R_{\alpha^{-1}}$
\index{transition} is the reciprocal relation of 
the relation $R_\alpha$;
thus it holds that $reg_w(b',\alpha) = prog(b', \alpha^{-1}) = \{s | R_\alpha(s) \cap b' \neq \emptyset\}$.


Postdiction \index{postdiction} must be distinguished from
{\em goal regression}, \index{goal regression} also called {\em strong regression},
which is the reverse transformation of progression. \index{progression}
It is defined only for the binary model \footnote{In the probabilistic model,
there may not exist a unique probability distribution
\index{probability} $b$ on $\ss$ satisfying $b'(s') = \sum_{s \in \ss}
b(s).p(s'|s,\alpha)$, $b'(s')$ with $p(s'|s,\alpha)$ being known for all $s$,
$s'$ and $\alpha$.}: given a belief state $b' \subseteq \ss$ and a purely ontic action $\alpha$,
the aim is to find the belief state $b = reg_S (b', \alpha)$
such that $prog (b, \alpha) \subseteq b'$ and $b$ is maximum for
set inclusion; this belief state is the least informative state of belief
(thus the least conjectural) which guarantees that the execution of $\alpha$ in it led
to the goal $b'$.

Let us notice that $reg_S(b',\alpha) \subseteq reg_w(b',\alpha)$ with the particular case $reg_S(b',\alpha) =
reg_w(b',\alpha)$ when $\alpha$ is deterministic.


Progression \index{progression} and regression \index{regression} are two key processes of reasoning
for {\em planning} \index{planning} (see Chapter 10 of Volume 2), which consists in determining the actions to carry out to make evolve the
system as the agent wishes it (for example, get as close as possible to a
reference trajectory in the case of the supervision, \index{supervision} or to
reach a goal state \index{goal} in the case of classical planning).
On the other hand, postdiction \index{postdiction} has little interest for
planning itself (because if $b$ is obtained as a possible postdiction from $b'$
with action $\alpha$, it is not guaranteed that by carrying out the action
$\alpha$ one would again obtain the state $b'$, while strong regression guarantees it by definition).


\subsubsection{Prediction and Postdiction with Epistemic Actions}
\index{postdiction} \index{prediction}
The progression \index{progression} of a belief state by an epistemic action depends
on the nature of the reasoning process.  In the case of a 
supervision \index{supervision} process or a diagnosis\index{diagnosis}, 
the agent reasons online and thus has access to all the observations coming from
the actions {\em feedback} during its reasoning; thus it is enough to define the
progression of a belief state by an observation, which is related to {\em
belief revision} \index{revision} (see Chapter 14 of this volume). 
In the binary model, the progression \index{progression} of one
belief state $b \subseteq \ss$ by an observation $o$ after having
carried out the action $\alpha$ is 
$b \cap S(o)$, where $S(o) = \{s | o \in O_\alpha(s) \} $;
while in the Bayesian model, the revision of $b$ by $o$ is 
the probability distribution $b(. | S(O))$. \index{probability}


The {\em filtering} process consists in determining the new state of belief
$b'$, given an initial belief state $b$, an action $\alpha$, and an observation $o$ resulting
from the execution of $\alpha$. In the binary model, this new belief state is
simply $prog(b,\alpha) \cap S(o)$.
In the Bayesian model, the probability distribution
\index{probability} $b'$ obtained
after having carried out $\alpha$ and observing $o$ is
$b'(s') = \frac{p(o | s', \alpha).\sum_{s \in S}b(s).p(s' | s,\alpha)}
{\sum_{s'' \in S} (p(o |  s'', \alpha).\sum_{s\in S}  b(s). p(s'' | s,\alpha))}$;
it is the formula expressing the revision of the beliefs by
the {\it feedback} in the partially observable Markov decision processes \index{Markov decision
  process (MDP)} (see Chapter 10 of Volume 2).


In the case of a planning process, \index{planning} where the aim is to build an
off line plan and to reason on its effects, the progression
\index{progression} of a belief state by an epistemic action is in general not a
unique belief state, but a set of such states (one for each possible
observation, since the actual observation cannot be known off line). In the binary model,
$prog (b, \alpha) $ is the set of belief states $\{b \cap S(o)$~$|$~$b \cap
S(o) \neq \emptyset \}$ for $o$ varying in $\Omega$.
For sake of shortness, we do not give details on regression
by epistemic actions.



\subsubsection{Event Abduction}

The third type of reasoning \index{reasoning!temporal}
is {\em event abduction}. \index{abduction}
It concerns reasoning on the event which took place between two successive time points $t$ and $t+1$, starting from the
description of the
possible events and from the description of the belief states at time $t$ and $t+1$ \footnote{
A more complex abduction problem consists in reasoning not
only on the event which took place, but also on the system states at time points
$t$ and $t+1$, on which one wishes to obtain  more precise beliefs.
}.
If the event in question is exogenous, this reasoning is called {\em explanation} \index{explanation}.


As for planning, progression \index{progression} and goal regression \index{regression}
are two key processes for event abduction: in planning, one must choose the actions to be carried out to make the system evolve as desired starting from its current state; in event abduction,
the objective is to determine which event $\alpha$ led the system to evolve as it did between $t$ and $t+1$
(even if this evolution was not desirable).
In the binary model, to compute such $\alpha$ consists in searching among the possible events
those satisfying $b' \subseteq prog(b, \alpha)$ (or equivalently $b \supseteq reg_F(b', \alpha)$).


\subsubsection{Scenario Extrapolation}
More generally, these types of reasoning, \index{reasoning!temporal} that we
defined in a context where there is only one change stage (thus two time points), take place
in situations where the horizon is unspecified, and where the input information
is a complex {\em scenario} describing a partial trajectory of the system 
(at each time point, some information may be available about the
occurrence of an action and/or an observation). In the typical case where
no action was carried out and where the user wants to find the events 
(or more simply, the elementary changes) which occurred at
each time point, the process is called {\em extrapolation}. \index{extrapolation} 

Another situation is when one seeks to recognize some trajectories among
a set of reference trajectories in order to predict the
events that will occur and/or the states that
the system will reach; this process is called \emph{scenario monitoring} or
\emph{scenario recognition}. 

The sequence of observations can also contain action occurrences \cite{DupindeSaintCyr08,DeLe12,HuDe15} (scenari are also called narratives or histories) and the two previous tasks of completing or recognizing some trajectories can be done in this more complex context. These tasks involve both prediction, postdiction and event abduction in situation that can be pervaded with uncertainty (fallible knowledge, erroneous perception, exogenous actions, and failed actions).

\bigskip

A crucial aspect of the reasoning about change approaches in artificial intelligence is that they assign a prominent role to {\em inertia}
\index{inertia}: by default, the system tends to remain static, and the
changes other than those which are directly caused by action occurrences are
rare, 
this is why one seeks to minimize them.
This assumption is crucial if one wants to reason about action
in presence of uncertainty without losing too much information. Very often,
reasoning about change amounts to {\em minimize change}
\index{change!minimal}; we will come back on this subject when we will approach the languages for
reasoning about action. Indeed, according to the way actions are represented,
there exist numerous ways of carrying out the progression of a state or the regression of a
formula (encoding a set of states) by an action. Concerning the temporal or
dynamic logic representations,
progression \index{progression}
and regression \index{regression}
can be computed via some formula transformations (in particular, conjunction and forgetting).
The use of change minimization principles is often
proposed as a means to solve the frame problem 
\index{problem!frame}
(cf.\ Section \ref{volume-1-chapitre12-problemedecor}),
but it seems henceforth admitted that it
is rather necessary to set up processes which remove the solutions containing abnormal changes (not caused by actions) than processes which minimize them.

This idea to focus on abnormal changes rather than on maximising inertia is well in accordance with the approaches that reason on a world under continuous change where the agent should adapt its action model when a surprise occurs (discrepancy between what is observed and what was expected). This kind of research is more related to the domain of planning in a dynamic word and particularly in the context of goal driven autonomy agents (GDA) that must reason about partially observable domains with a partial knowledge about available actions \cite{MoAh14,DMC16,DaCo17}.

\section{Reasoning about Action:  Languages} \label{volume-1-chapitre12-languages}

\subsection{Problems related to the Representation of Actions} \label{volume-1-chapitre12-problemedecor}

In the majority of real problems, the system is naturally
described by some variables, called \emph{state variables}, that represent some
features about some objects, etc.  In this case, a state of the system corresponds to the description of a value for each one of the state variables, these values being able to change with the course of time. State variables are usually called {\em fluents}, a fluent describes
a dynamic property of the system. Obviously, the number of possible states
is exponential in the number of variables. The {\em explicit} description
of the effects of the actions, which consists in specifying {\em in extenso}
the functions $R_{\alpha}$, becomes then unfeasible in practice, and is somewhat unnatural,
because the user is obliged to describe the actions state by state. 
Similar considerations apply to the description of the correlations
between states and observations and with the computation of the operations of progression, \index{progression}
regression, \index{regression} etc.

However, there often exist much more economic and natural ways to represent the effects of the actions.
For example, let us consider the action to ``flip a switch''
which causes the alternative ``lighting on'' or ``off'' of a bulb.
If the representation of the problem requires to take into account the ``on''/``off'' states of 10 bulbs, then 1024 states of the system  will have to be considered
(all possible configurations of the 10 bulbs) in order to
describe one flip action, whereas this action only causes the change of state
of one particular bulb. To describe such
action, one rather wants to be limited to indicate that it changes the state of this bulb and, implicitly, that it
leaves the other bulbs in their current state.

\emph{Action languages} \index{language!action} were built
precisely to this aim: obtaining representations of the effects of the actions which are both more economic (or more compact) and more natural.
The problem of preventing the user from explicitly describing the fluents that are
not modified by an action in the various possible contexts is known as 
{\em the frame problem}
\index{problem!frame}
\cite{McHa69}.
It is indeed a problem involved in the choice of a representation of the actions
(and not of a modeling problem, i.e. the problem
does not rely on the choices of the fluents used to model the system but on the coding
of actions in general).

In the same vein, one may face a problem that is the dual of the frame problem, known as {\em ramification problem}
\index{problem!ramification} \cite{Finger87} which is solved when the action
language makes it possible to avoid describing 
explicitly all the fluents that an action modifies, directly or indirectly, in
the various possible contexts. Following up on the previous example, each flip of
the switch causes the lighting on/off of the associated bulb, then the room where
the bulb is located becomes enlightened and 
consequently one can settle there for reading. This derived fact is a consequence of the execution of the action but
it is not natural, when the action is described, to specify it directly: it
results rather from a (static) law which expresses that when a room is lit, one can practice the reading there.

When one deal with action representation, the {\em qualification}
\index{problem!qualification} problem \cite{McCarthy77} is also often evoked; this problem
expresses the incapacity to describe all the pre-conditions that guarantee to obtain
the ``normal'' effect of an action. To deal with this problem, it is first necessary to circumscribe the
world with the individuals and the properties explicitly present in the
representation; for example, 
flipping the switch 
when the associated bulb is off will cause the lighting of this one only if the
conflict between Bordures and Syldaves did not cause the destruction of the
electric line feeding the house. From our point of view, this problem is not
intrinsic within the action representation, it occurs more primarily as soon as
the modeling phase starts and simply reflects the
difference existing between a situation of the physical world and a
representation of this one, which necessarily abstracts it. However, in order
to give the pre-conditions of an action, this
restriction to the situations that have a representation in the language does not remove
the need for reviewing all the situations in which the action is carried out
normally. Solving the qualification problem
\index{problem!qualification} means being able
to state the ``natural'' pre-conditions of an action without having to describe
explicitly the list of all the values of the fluents which allow the action to
normally take place.



Once actions are represented, it is necessary to build
algorithms allowing the computation of the basic operations
(progression, \index{progression} regression, \index{regression} etc).
The choice of an action language thus depends, on the one hand, of its more or
less natural aspect, on the other hand, of its compactness (or space
efficiency), and finally, of the complexity of the basic operations when the
actions are represented in this language (its computational efficiency). 

There exist many action languages which were developed and studied by the community.
They can be gathered in several families, according to the nature of the mathematical objects
that they use (propositional or first order logic formulas, temporal or
dynamic logic formulas, Bayesian networks, state automatons, etc).
Giving an exhaustive panorama would be too long and little digest.
We will thus only sketchily present 
the languages which received the most attention from the community, and
which are sufficiently representative of the range of the existing languages.
Each following sub-section approaches a particular language, or a family of languages,
by briefly giving its specificities.

\subsection{The Situation Calculus}\label{volume-1-chapitre12-sitcalc}

From an historic perspective the \emph{situation calculus} \index{situation calculus} introduced
by McCarthy and Hayes \shortcite{McHa69} is the first formalism devoted to reasoning about actions.
The definitions given by these authors enabled them to set the basic concepts (presented higher)
on reasoning about change and action.
The situation calculus is a typed first order logic \index{logic!first order} language \index{language!action} 
with equality, whose types are fluents, states (called situations),
actions and objects.
In order to simplify the presentation, here we only consider 
propositional fluents, which have one situation as single argument;
we do not mention the objects of the world.
Thus, $\lnot P(S_0)$ express that the fluent $P$ is false in the situation $S_0$.
$S_0$ denotes the state of the system at the initial time point of the horizon.
For situations and actions we need both variables
(denoted respectively $s, \ldots$ and $x, \ldots$)
and constants
(denoted respectively $S, \ldots$ and $A, \ldots$).
The function $do$ applies to a situation and an action and returns a situation.
Thus, the formula $\lnot P(S_0) \wedge P (do(A_1, S_0))$
expresses that $P$ is false in $S_0$ and true in $do (A_1, S_0) $, i.e.
in the situation obtained by applying the action $A_1$ in $S_0$.
The formula $\forall s \lnot P(s)$ expresses that $P$ is always false.
The formula $\forall s ((\forall x \lnot P(do(x,s)) \leftrightarrow x = A_0)$
expresses that $A_0$ is the single action which guarantees to make $P$ false in any state where it is applied.

McCarthy and Hayes set a general representation framework 
enabling them to represent actions by their pre-conditions and their effects
(represented by logical formulas).
Many approaches were then proposed in order to characterize the ``good''
consequences of these formulas.
Initially, all the authors bet on {\em change minimization} in order to
restrict the set of models so that the properties resulting from the inertia
principle can be deduced without having to mention them explicitly.
This was accomplished thanks to a second order logic formula,
and various circumscription policies  \index{circumscription} were studied for this purpose
(the reader can refer to \cite{Moinard00} for a review).
McCarthy \shortcite{McCarthy86} and then Hanks and McDermott
\shortcite{HaMD86} used the circumscription of abnormality predicates (by
considering that a fluent must persist unless otherwise explicitly indicated) 
within the framework of the situation calculus.
However, there are some examples where circumscription does not give the
expected result. 
One of most famous is the {\em Yale Shooting Problem} proposed by Hanks and
McDermott: someone is alive in the initial situation, and one
carry out successively the three actions ``Load'', ``Wait'' then ``Shoot''.
The action ``Shoot'' is described by the formula: $\forall s$, (loaded ($s$) $\rightarrow$
(Abnormal (Alive, Shoot, $s$) $\wedge$ $\neg$Alive (do(Shoot, $s$))))\footnote{If the
rifle is loaded in the situation $s$ then the fluent ``Alive'' is abnormal
(i.e., non persistent) when the action ``Shoot'' takes place in $s$ and the person will not be alive any more in
the resulting situation.}. The fact
that, by default, the fluents are persisting is described by the second order
logic formula 
$\forall f$, $s$, $a$, (($f(s)$ $\wedge$ $\neg$Abnormal ($f$, $a$, $s$) $\rightarrow$
$f$ (do($a$, $s$))\footnote {If the fluent is not abnormal with respect to an
  action then it keeps its value after the execution of this action.}.
The circumscription of the predicate
$Abnormal$ makes it possible to obtain a logical model in which the person is alive at
the initial time point and dead (non alive) after the action ``Shoot''. However, another
model is possible: the one where the rifle unloaded itself during ``Waiting'' and
the person is still alive after ``Shoot''. Circumscribing the
$Abnormal$ predicate \index{circumscription}
\index{predicate}
does not allow for preferring the first model to the second one because the two models have
incomparable sets of abnormalities w.r.t. set inclusion (in the first model, it is ``Alive'' which is abnormal
in the presence of the action ``Shoot''; in the second one, it is ``Loaded'' which is abnormal w.r.t. 
``Wait''). Chronological ignorance,
proposed by Shoham \shortcite{Shoham88} and consisting in preferring models where
the changes occur the latest, allows one to obtain a satisfactory answer for this
example. 
But this last {\it ad hoc} approach  was challenged by
other examples 
that it handles badly \cite{Sandewall95, FrHa94}.



Another solution suggested by Lifschitz and Rabinov \shortcite{LiRa89} is to impose that all
the fluents that are modified by an action are systematically non inert
when this action is carried out. This idea is close to the solution,
proposed by Castilho, Gasquet and Herzig \shortcite{CGH99}, to use a
dependence relation between an action and the atoms on which it may act.
The reader can refer to \cite{Sandewall95} for an excellent synthesis of all these works.

In short, approaches based on change minimization 
are based on {\em non-monotonic} logics
\index{logic!non-monotonic} and are very complex;
they are not able to deduce all the intuitive consequences
that are expected from a description of a set of actions and an initial situation.

\bigskip

The situation changed with the publication of what was called
{\em Reiter's solution} to the frame problem \cite{Reiter91}.
\index{problem!frame}
Reiter suggests a \emph{monotonic} solution \index{monotony} based on 
{\em successor state axioms} (SSA).
These axioms must be given for each fluent $P$
(which is equivalent to an assumption of complete information about the conditions
of change of truth value of a fluent)
and they have the following form:
$$\forall s,x \quad ( P(do(x,s)) \leftrightarrow \gamma_P(x,s) ) $$
where $\gamma_P(x,s)$ is a formula which does not contain the function symbol $do$
and which can only contain $S_0$ as situation constant.
Thus, the SSA for $P$ describes the conditions under which
$P$ is true after an action has been performed, in function of what was true before.

Let us consider the Toggle-switch example  \cite{Lifschitz90}: \\
\parbox{0.7\textwidth} {In a room, the light is on only if both
switches are up or both down.
Initially, the switch $a$ is up and the switch $b$ in down, the light
is thus off, someone toggles the switch $a$.} \hfill
~ \parbox{0.3\textwidth}{\includegraphics[width=0.3\textwidth]{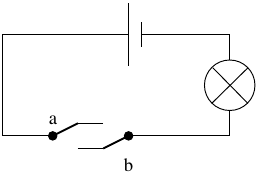}}
The fluents are $U_a$ (``the switch $a$ is up'') and $U_b$ (``the switch $b$ is up'').
In this example, the SSA for fluent $U_a$ can be written:
$$\forall s, x \quad  U_a(do(x,s)) \leftrightarrow  ( (\lnot U_a(s) \land x =    T_a ) \lor
                                                    (      U_a(s) \land x \neq
 T_a ))$$
where $T_a$ is the action 
to toggle the switch $a$, i.e., flip its position.

Reiter explains that using Successor States Axioms is a solution to the frame problem
\index{problem!frame}
because one can reasonably expect the size of the set of SSA
to be in the order of the number of fluents 
(which contrasts with the size of the explicit description of the frame axioms
that would be in the order of the number of fluents set multiplied by the number  
of actions).


According to Reiter, quantification over actions is the key solution to the
frame problem. \index{problem!frame}
As we will show in Section \ref{volume-1-chapitre12-pdl}, the assumption of complete information about the conditions under which fluents change their truth value (translated by the $\leftrightarrow$ in the SSA)
allows Reiter to deal with the frame problem in a satisfactory way.


The presence of an SSA for each fluent allows {\em for regressing} formulas \index{regression}:
atoms of the form $P(do(\alpha,\sigma))$
(where $\alpha$ and $\sigma$ are terms built with variables, constants
and the function $do$)
are replaced by the right member of the SSA for $P$,
by applying first the suitable substitution \index{substitution};
this process is reiterated until complete elimination of the function $do$.
By construction, the formula thus obtained only relates on the initial state $S_0$.


For example, the formula
 \\ \centerline{$U_a(do(T_a, do(T_a,S_0)))$}
is first replaced by:
\\ \centerline{
$(\lnot U_a(do(T_a,S_0)) \land T_a {=}    T_a ) \lor
 (      U_a(do(T_a,S_0)) \land T_a {\neq} T_a ) $
 }
which can be simplified into
$\lnot U_a(do(T_a,S_0)) $.
In a second step, this last formula is replaced by
$\lnot
 (\lnot U_a(S_0) \land T_a {=}    T_a ) \lor
 (      U_a(S_0) \land T_a {\neq} T_a ) $
which can be simplified into
$U_a(S_0) $.
We have thus proven by regression \index{regression}
that the switch is up after two executions of $T_a$ if and only if it is up
in the initial state $S_0$.


In order to decide whether the application of the action $\alpha$ in the state $S_0$ leads to a state in which
$\psi$ holds, it is enough to decide if the formula
$\phi(S_0) \rightarrow \psi (do (\alpha, S_0))$ is valid.
The regression of $\psi (do (\alpha, S_0))$ results in a formula
$\psi'(S_0)$. \index{regression}
If the argument $S_0$ is eliminated, we obtain the propositional formula 
$\phi \rightarrow \psi'$,
whose validity can be checked by using a suitable prover.

This solution was combined with epistemic logic
\index{logic!epistemic}\cite{ScherlLevesque03},
which gives a formalism close to dynamic logic,
described in Section \ref{volume-1-chapitre12-pdl}. Moreover the framework of the situation calculus with a Successor State Axioms has been recently used by \cite{BaSo18} for causal ascription.

\subsection{Propositional Action Languages}\label{volume-1-chapitre12-propositionnel}

A weak point of the approaches based on the situation calculus is the difficulty of their algorithmic implementation.
For this reason, researchers have also developed approaches based on propositional logic, which can benefit from
off-the-shelf ASP or SAT solvers  \index{satisfiability of Boolean formulas (SAT)} \index{solver!ASP} 
(see Chapters 4 and 5 of Volume 2).


In action languages based on propositional logic, action effects are represented by
local rules specifying only the fluents that change, possibly together with the conditions
under which they change. Let $F$ be a finite set of fluents. The states of $\ss$ are the 
propositional interpretations\index{interpretation}\index{logic!propositional} 
over $F$, that is, $\ss= 2^F$.


The most basic action language\index{language!action} is arguably \emph{STRIPS}\index{STRIPS} \cite{FiNi71}, 
where an action is represented by a precondition and its effects (see Chapter 10 of Volume 2),
a  precondition being a conjunction of literals and an effect being a consistent set of literals.\index{planning}


To encode the light switch example, one may take as set of fluents
$F = \{U_a,U_b\}$, where $U_a$ (resp. $U_b$) is true (resp. false) when switch $a$
(resp. $b$) is on (resp. off). An action with conditional effects such as 
$T_a$ (`switch $a$') can be written as
$$(U_a \mapsto \neg U_a) \wedge (\neg U_a \mapsto U_a).$$
The right member $l$ of each rule of form $c \mapsto l$ is a direct action effect,
which applies if and only if the corresponding condition is satisfied in the state in which
the action is performed. Thus, applying $c \mapsto l$ in state $s$ leaves $s$
unchanged if $s$ does not satisfy $c$ and enforces the truth of $l$ otherwise, leaving other fluents 
unchanged. This applies to each rule\footnote{A pathological case is when the conditions of rules
leading to complementary literals are conjointly satisfied in $s$; in such a case, the 
progression \index{progression} is undefined; this can reflect an error when specifying the 
representation of the action, or the fact that $s$ is impossible (and in this case corresponds to
an implicit static law).}. Thus, applying $T_a$ in state $s$ leads to change the truth value of
$U_a$ in $s$, as we expect. Importantly, such an action description rule is not a classical logical
formula, and in particular,  $\mapsto$ is not material implication. Indeed, a STRIPS \index{STRIPS} 
action $\alpha$ can be seen as a {\em constraint} linking the state of the world {\em before} it is performed and
the state of the world {\em after} it has been performed. In particular,
$c \mapsto l$ is not equivalent to $\neg l \mapsto \neg c$.\footnote{If they were equivalent,
then the encoding of action ``Shoot'' by \emph{Loaded} $\mapsto \neg$ \emph{Alive} in the
{\it Yale Shooting Problem} would be equivalent to  \emph{Alive} $\mapsto \neg$ \emph{Loaded}, 
meaning that shooting on a living person results on the gun being magically unloaded (and the person staying alive...)}

One of the limits of the STRIPS language is the impossibility to express static laws. These laws are however
needed for the ramification problem to be dealt with\index{problem!ramification}. 
For instance, in the previous example, one may want to introduce a new fluent
$L$ expressing that ``the light is on''. With standard STRIPS, integrating this new
fluent would require to modify all actions by specifying what happens to $L$.
This solution is not reasonable when the number of fluents is large. A way to
cope with this lack of expressiveness consists in encoding actions with a set
of {\em basic} fluents on which the available actions act directly 
($U_a$ and $U_b$ in the example); the fluents that are not basic are called {\em derived} fluents. 
Progression is first computed as in classical STRIPS, and then there is one additional step so as to take
the static laws into account and make some inferences on derived fluents. 
Thus, to compute the progression \index{progression} of state $s$ by an action,
one starts by projecting $s$ on the basic fluents; then one performs the progression of
this projection, and finally the obtained state is completed using the static laws. In the
switch example, one may take as static law 
$$((U_a \wedge U_b) \vee (\lnot U_a \wedge \lnot U_b)) \leftrightarrow L$$
where $U_a$ and $U_b$ are basic and $L$ is derived. The 
progression\index{progression} of state $\{U_a,\neg U_b, \neg L\}$ by action $T_a$ is thus $\{\neg U_a,\neg U_b, L\}$.

There are four main problems with STRIPS \index{STRIPS}: it does not allow for representing 
(a) non-determinism,
(b) static causal relations between fluents (as discussed in the previous paragraph),
(c) concurrent actions, and
(d) epistemic actions. To cope with this lack of expressiveness,
more sophisticated action languages\index{language!action} have been developed, 
both in the planning community (with ADL \cite{Pednault89} and PDDL \index{PDDL} \cite{PDDL98})
and in the knowledge representation community. We will now focus on the languages stemming
from the latter community.

In the 70s and 80s, the knowledge representation community used to think of actions as simple rules linking
action preconditions and action effects. 
Subsequently, some researchers suggested that prediction\index{prediction} 
could be computed using \emph{minimization of change}, so as to impose that, by default, fluents that are not concerned
by the action should persist (these fluents, of course, do not need to be specified in action effects, so as to cope with the
frame problem).
\index{problem!frame}
Then, since the 90's, minimization of change was progressively replaced by the use of propositional languages based on
{\em causal implication}. The solutions of \cite{Reiter91}, \cite{LiRa89} and \cite{CGH99} for solving problems occurring with
minimization of change consist in expressing dependencies\index{dependency} between an action and
its effects. This very principle has been implemented in works using \emph{causal implication}\index{causality} (see Chapter 9 of this volume), 
which is distinct from material implication since it is meant to express these dependencies.

Some approaches using causal implication make use of the situation calculus
 \cite{StMo94}, \cite{Lin95}. Others use the modality C \cite{Geffner90,GMS98,Turner99} 
 or equivalently, define a new connective $\Rightarrow$ \cite{GLLMT04}. Yet others
define influence relations between fluents  \cite{Thielscher97}. The main feature of
these approaches is that they distinguish the fact of being true from the reason for
being true, and use this distinction for computing the expected effect of actions for
prediction \index{prediction} or planning. \index{planning}


We give now some details about the action language \index{language!action} ${\cal A}$ 
proposed by Gelfond and Lifschitz \shortcite{GeLi93}. In this language, an action is expressed by
means of {\em conditional causal rules} of the form 
\begin{center}
{\sf if $c$ then $\alpha$ {\sf causes} $l$},
\end{center}
where
$\alpha$ is an action name, $c$ a conjunction of literals (omitted when it is equivalent to $\top$),  
and $l$ a literal\index{literal}. A set of causal rules defines a deterministic transition system 
\index{transition} between states. Thus, the action $\alpha$ defined by the causal rules 
\index{rule!causal} 
\begin{center}
{\sf if $p \wedge q$ then $\alpha$ {\sf causes} $\neg p$,  
if $\neg p \wedge q$ then $\alpha$ {\sf cause} $p$ and $\alpha$ {\sf cause} $q$}
\end{center}
corresponds to the
transition system $R_\alpha$ defined by $R_\alpha(pq) = R_\alpha(\bar{p}\bar{q}) = \bar{p}q$ and
$R_\alpha(\bar{p}q) = R_\alpha(p\bar{q}) = pq$. An action $\alpha$ described by such causal rules
corresponds to a {\em propositional action theory} $\Sigma_\alpha$, expressing $\alpha$
by means of propositional symbols $F_t$ and $F_{t+1}$, with $F_t = \{f_t | f \in F\}$ and 
$F_{t+1}= \{f_{t+1} | f \in F\}$, where $f_t$ represents fluent $f$ at time
$t$,\index{logic!temporal}
that is, before action $\alpha$ has been performed, and $f_{t+1}$ represents $f$ at time $t+1$,
after action $\alpha$ has been performed. The causal rules are translated into $\Sigma_\alpha$ 
according to the following principle: fluent $f$ is true at $t+1$ if and only if one of these two
conditions holds: (a) it was true at $t$ and the state at $t$ does not satisfy any precondition of a causal
rule whose conclusion is $\neg f$, or (b) it was false at $t$ and the state at $t$ satisfies the 
precondition of a causal rule whose conclusion is $f$. One finds here the principle at work in
the situation calculus, which we called `Reiter's solution' in Section \ref{volume-1-chapitre12-sitcalc}.

Formally, let $\Gamma(f)$ (respectively $\Gamma(\neg f)$)
the disjunction of all preconditions of rules whose
conclusion is $f$  (respectively $\neg f$); then
$\Sigma_\alpha$ is the conjunction of all the formulas $$f_{t+1}
\leftrightarrow \Gamma(f)_t \vee (f_t \wedge \neg \Gamma(\neg f)_t)$$ for
$f \in F$.
Thus, the action theory $\Sigma_\alpha$ corresponding to the action $\alpha$ 
previously described by its causal rules \index{rule!causal}
is $$\Sigma_\alpha = (p_{t+1} \leftrightarrow ((\neg
p_t \wedge q_t) \vee (p_t \wedge \neg (p_t \wedge q_t)))) \wedge
(q_{t+1} \leftrightarrow \top),$$
 which simplifies into $$\Sigma_\alpha = q_{t+1} \wedge (p_{t+1} \leftrightarrow (p_t \leftrightarrow \neg
  q_t)).$$
  

An extension of language ${\cal A}$ is language \index{language!action} ${\cal C}$ \cite{giu98},
which allows for expressing executability conditions and static rules, independently of any action, such as
\begin{center}
{\sf \small Outside} $\wedge$ $\neg${\sf \small Umbrella} $\wedge$ {\sf \small Umbrella} {\sf causes} $\neg${\sf \small Dry},
\end{center}
that are also taken into account in the corresponding action theory.
For instance, consider action {\sf \small Go-out} with a unique causal rule  
\begin{center}
{\sf \small Go-out} {\sf causes} {\sf \small Outside};
\end{center}
\index{rule!causal}
the corresponding action theory, taking 
into account the previous static rule, is 
\begin{center}
\begin{tabular}{ll}
$\Sigma_{\sf Go-out}$ = & \sf Outside$_{t+1}$ $\wedge$ (Umbrella$_{t+1}$ $\leftrightarrow$ Umbrella$_t$)
$\wedge$ (\sf Rain$_{t+1}$ $\leftrightarrow$ Rain$_t$) \\ & $\wedge$ \sf
(Dry$_{t+1}$ $\leftrightarrow$ Dry$_t$ $\wedge$ (Umbrella$_t$ $\vee$ $\neg$Rain$_t$)).
\end{tabular}
\end{center}


{\em Non-determinism} can be expressed in several different ways, explored independently in different papers:
\begin{itemize}
\item by {\em complex effects}, such as 
\begin{center}
$\alpha$ {\sf causes} $(p \leftrightarrow q)$,
\end{center}
a choice that is
at the heart of belief update, cf.\ Section \ref{volume-1-chapitre12-sec:update}; 
\item by {\em disjunction of effects}, which are similar to nondeterministic union in dynamic logic, cf.
Subsection \ref{volume-1-chapitre12-pdl}), such as 
\begin{center}
{\sf \small Toss} {\sf causes} {\sf \small Heads} {\sf or}  {\sf causes} $\neg${\sf \small Heads}; 
\end{center}
\item by recursive causal rules, which is a more technical solution that we will not discuss here.
\end{itemize}
Some action languages\index{language!action} (such as language
${\cal C}$) also have {\em concurrency}, whereas others have
{\em epistemic actions}, thus enabling the distinction between facts and knowledge; thus, the action of testing whether the 
fluent $f$ is true or false is represented by the causal rule 
\index{rule!causal}
\begin{center}
$\alpha$ {\sf causes} {\bf K}$f$ {\sf or causes}  {\bf K}$\neg f$,
\end{center}
where {\bf K} is the knowledge modality \index{knowledge} 
of epistemic logic \index{logic!epistemic}{\sf S5} (see in particular \cite{HerzigLangMarquis03}).


{\em Progression} \index{progression} and {\em regression} \index{regression} can be applied directly
in these languages. A belief state, in the binary uncertainty model\index{uncertainty}, is a nonempty set of states, and can thus be
represented by a consistent propositional formula. Progression and regression map a consistent formula and an action to a formula (which is always consistent in the case of progression and weak regression).  The progression of formula $\varphi$ by action $\alpha$ consists first in taking the conjunction of $\varphi_t$  (expressing that $\varphi$ is true before the action) and $\Sigma_\alpha$, and then in forgetting in $\varphi_t \wedge \Sigma_\alpha$ all variables $f_t$, {\em i.e.}, in deriving the strongest logical consequence of  $\varphi_t \wedge \Sigma_\alpha$ 
independent \index{independence} of the variables $f_t$ (see for instance \cite{LangLM03}).
Weak regression \index{regression} is computed similarly: the weak regression of $\psi$ by $\alpha$ is the result of forgetting the
variables $f_{t+1}$ in $\psi_{t+1} \wedge \Sigma_\alpha$. The strong regression of $\psi$ by $\alpha$ is obtained by computing the minimal conditions guaranteeing that the application of $\alpha$ will lead to a state satisfying $\psi$. Thus, in the previous example, the progression of {\sf \small Dry $\wedge$ Umbrella} by {\sf \small Go-out} is (up to logical equivalence) 
\begin{center}
{\sf \small Outside} $\wedge$ {\sf \small Umbrella} $\wedge$ {\sf \small Dry},
\end{center}
and the progression of {\sf \small $\neg$Umbrella} by {\sf \small Go-out} is 
\begin{center}
{\sf \small
Outside} $\wedge$ $\neg${\sf \small Umbrella} $\wedge$ ({\sf \small Rain}
$\rightarrow$ $\neg${\sf \small Dry}),
\end{center}
whereas the weak regression  of {\sf \small Dry} $\wedge$ {\sf \small Rain} by {\sf \small Go-out} is 
\begin{center}
{\sf \small Umbrella} $\wedge$ {\sf \small Rain} $\wedge$ {\sf \small Dry}.
\end{center}
\subsection{Dynamic Logic}\label{volume-1-chapitre12-pdl}

There are other possible ways of representing actions and dealing with the corresponding problems.
\emph{Dynamic logic}\index{logic!dynamic} is a formalism initially known in theoretical 
computer science for reasoning about program execution.\index{program}
In addition to Boolean operators, its language contains {\em modal operators}\index{logic!modal}
of the form $[\alpha] $, where $\alpha$ is a program.
The combination of such an operator with a formula results in a formula of the form
$[\alpha] \phi$, read `$\phi$ is true after every execution of $\alpha$'.
Instead of a program, one may assume that $\alpha$ is an event or an action.
For instance, the action of toggling switch $a$ can be described by the two effect laws
$(\lnot U_a \rightarrow [T_a]       U_a )$ and $( U_a \rightarrow [T_a] \lnot U_a )$.


In the context of dynamic logic, an important aspect of reasoning about actions that was dealt with first in \cite{HerzigVarzinczak-Aij07} concerns the consistency of a domain description.
It has been shown that for expressive action languages\index{language!action}, beyond logical 
consistency, a good domain description should be modular, in the sense that effect laws describing
the actions should not allow for deriving new static laws. For instance, the static laws
$P_1 \rightarrow [A]Q$, $P_2 \rightarrow [A]\lnot Q$ and $\lnot [A]\bot$
together imply the static law $\lnot (P_1 \land P_2)$; if this law is not deductible from the
other static laws and only them, then these effect laws should be considered problematic.


Unlike in situation calculus, states are not explicit in dynamic logic.
Although dynamic logic does not allow either for quantifying over actions, 
which is a key feature or Reiter's solution for the frame problem, it has been
shown that this solution can be implemented in dynamic logic\index{logic!dynamic} 
for the rather general case of explicit SSAs \cite{DitmarschHerzigLima-Jlc10}.
In such SSAs, $x$ must be the only action variable of $\gamma_P(x,s)$
and if an action constant $A$ does not appear in $\gamma_P(x,s)$ then
$\gamma_P(A,s)$ should not be equivalent to $P(s)$.
These conditions are natural for a system satisfying inertia. An example of
SSA not satisfying them would be
$\forall s (\forall x P(do(x,s)))\leftrightarrow \lnot P(s))$,
which means that $P$ is changed in every state (thus $P$ is a non-inert
fluent). Note that the formula $\gamma_{U_a}(x,s)$ in our
example from paragraph \ref{volume-1-chapitre12-sitcalc} satisfies these conditions.
In order to translate these SSAs in dynamic logic, one introduces {\em assignment actions} of the form
$P := \phi$; such an assignment describes an action where $P$ takes the truth value that $\phi$ had in the
previous state. This allows for associating with each action constant $A$ the following set of assignments:
$$
\sigma_{SSA}(A)\ \ =\ \
\{ P :=  simp( \gamma_P(A) )  \ \mid\ P \mbox{ appears in } \gamma_P(x) \}
$$
where $simp(\gamma_P(A) ) $ is obtained from $\gamma_P(x)$ by eliminating
argument $s$, substituting $x$ by $A$ and simplifying the equalities. 
In our example from paragraph \ref{volume-1-chapitre12-sitcalc}, 
after substituting
\index{substitution} $x$ by $T_a$ we obtain:
\\ \centerline{
$\sigma_{SSA}(T_a)\ \ =\ \
\{ U_a := ( \lnot U_a \land T_a {=}    T_a ) \lor
          (       U_a \land T_a {\neq} T_a )
\}$
}
which can then be simplified into
\\ \centerline{
$\sigma_{SSA}(T_a)\ \ =\ \
\{ U_a := \lnot U_a \}$.
}

Each occurrence of an abstract action symbol $A$ is replaced by the corresponding assignment.
As shown in \cite{DitmarschHerzigLima-Jlc10}, this constitutes a solution (in Reiter's sense) to the frame problem\index{problem!frame}. 
Thus Reiter's solution is
transferred to dynamic logic, without any need to quantify over actions. 
It is also shown that Reiter's 
solution can be combined with epistemic logic\index{logic!epistemic},
thus bridging it with epistemico-dynamic logics\index{logic!epistemico-dynamic}
(see Chapter 2 of this volume).

A recent work in dynamic logic is to investigate
epistemic extensions that are suitable for conformant planning \cite{LiYW17} 
and more generally for multiagent epistemic planning \cite{AucherB13,BolanderJS15,CooperHMMR16}.
An overview paper about combinations of logics of action with logics of knowledge and belief is \cite{Herzig15}.


\subsection{Dynamic Bayesian Networks}\label{volume-1-chapitre12-tbn}

\begin{figure}
\centering
 \parbox{0.4\textwidth}{\includegraphics[width=0.4\textwidth]{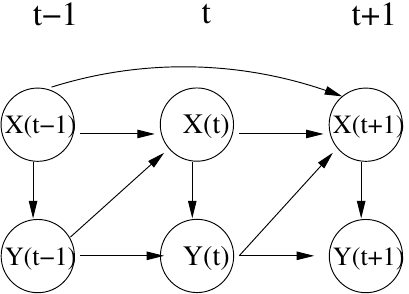}}
\caption{The DAG of a dynamic Bayesian network}
\label{volume-1-chapitre12-rbd}
\end{figure}

A \emph{dynamic Bayesian network}\index{Bayesian network!dynamic} \cite{DeKa89}
is a Bayesian network  (see Chapter 8 of Volume 2) in which the variables exist in as many copies as there are time points: for any fluent $f$ and time step $t$ there is a fluent $f_t$.
For each time step $t$, there exists a Bayesian network linking the variables
corresponding to $t$. Moreover, between these `instantaneous' networks, the only allowed
edges are those that are directed from past to future. The temporal directed acyclic graph (DAG)\index{graph}
given on Figure \ref{volume-1-chapitre12-rbd}, equipped with
probabilities \index{probability} for each variable, at each time step,
conditionally on the values of the parents of the variable, constitutes a dynamic Bayesian network.


If the system is Markovian, in order to determine completely the behavior of the system it is enough to
know the probability distribution for $x_t$ and the conditional probability distribution for $x_{t+1}$ given $x_t$.
The Markovian assumption can reasonably be made for many classes of systems. A Markovian temporal DAG
cannot admit an arc linking variables distant from more than one time step\index{time}: by deleting the edge
between $x_{t-1}$ and $x_{t+1}$ the diagram on the example below becomes Markovian, and a description restricted
to time steps $t$ and $t+1$ suffices.


The truth value of a fluent $f$ at a given time step can depend on its value at earlier time steps
$t-\Delta$, which translates in probabilistic terms into the following ``survival equation'':
$$Pr(f_t)=Pr(f_t\;|\;f_{t-\Delta}).Pr(f_{t-\Delta}) +
Pr(f_t\; |\;\neg f_{t-\Delta}).Pr(\neg f_{t-\Delta}).$$
The conditional probability \index{probability} $Pr(f_t\;|\;f_{t-\Delta})$ is called 
\emph{survival function}.
The survival function represents the tendency of propositions to persist given all events that can make them false.
A classical survival function is:
$Pr(f_t\;|\;f_{t-\Delta})=\exp^{-\lambda.\Delta}$, which indicates that the probability that
$f$ persists decreases, from the last time step where $f$ was observed to hold, at an exponential speed
determined by $\lambda$.
%

If one has some information about events that can affect the truth value of the fluent, then the survival equation no longer fits.
Generally, the probability that a proposition $f$ is true in $t$ is a fonction of:
\begin{itemize}
\item the probability $Pr(f_{t-\Delta})$ that it is true at $t-\Delta$ 
\item the probability $Pr(do(f_t))$ of the occurrence of an event that makes $f$ true at $t$
\item and the probability $Pr(do(\neg f_t))$ of the occurrence of an event that makes $f$ false at $t$.
\end{itemize}


From the standpoint of expressiveness, the interest of this class of probabilistic approaches
\index{probability} for reasoning about change is that it allows for expressing 
numerical uncertainty on beliefs,  observations, and causal laws (see also  \cite{HaMD94} and by
 \cite{Pearl88}).
On the other hand, a problem is that it requires the specification of many prior probabilities, even if it is not always necessary to
`solve' the whole probabilistic network to determine the probability of a proposition: one can
instead focus on a few key time steps (and key propositions). Note that the use of a dynamic possibilistic
network \index{logic!possibilistic}\index{network!possibilistic}
\index{possibility} allows one to reason without knowing precisely these probabilities
 \cite{HBABA07}. 


\section{Reasoning about Change: Update}\label{volume-1-chapitre12-sec:update}

\emph{Update}\index{update} is a research domain at the intersection of reasoning about actions (whence its presence in this chapter) and belief change 
(see Chapter 14 of this volume).  
It is a process for integrating into a belief base \index{belief base} 
a modification \index{belief change} of the state of the system that is explicitly specified by a propositional formula. 
More precisely, given a belief base $K$\index{epistemic state} 
and a propositional formula $\alpha$\index{propositional logic}, 
the update of $K$ by $\alpha$ is the progression \index{progression} of $K$ by an action,
or alternatively by an exogenous event whose occurrence is known and whose effect is $\alpha$.
Belief update is opposed to belief revision \index{revision} where
a new piece of information about the system is integrated into a belief base about that system,
under the hypothesis that the latter did not evolve.
The distinction was clarified by Katsuno and Mendelzon \shortcite{KaMe91}, 
although update was studied before \cite{KeWi85,Winslett88}, 
partially by scholars from the database community (see Chapter 3 of Volume 3).

The distinction between revision and update deserves to be clarified a bit more here.
If the new piece of information (`the input') completes our beliefs about the world then
it is not the world that has evolved, but only the agent's beliefs about the world. 
(This may be due to the questioning of an erroneous information about the world
or a new piece of information about the characteristics of the world.) 
In that case we have to perform a 
\emph{revision}\index{revision}.\footnote{
As shown in \cite{FriedmanHalpern99}, revision remains relevant even if the initial belief state and the new formula do not refer to the same time point, as long as there is a syntactical distinction (via some time-stamping) between a fluent at a time point and the same fluent at another time point:
what matters for revision is not that the world is static, but that the propositions that are used to describe the world are static.
This also explains that belief extrapolation also corresponds to a revision process \cite{DupinLang11}.
}
Such a revision amounts to a simple addition (also called {\em expansion}\index{expansion}) when 
the input is consistent with the beliefs; however, in case of inconsistency\index{inconsistency} 
revision selects some beliefs that have to be rejected in order to restore consistence\index{consistence!restoring consistence} 
(see Chapter 14 of this volume).
If the input characterizes an explicit evolution of the world (i.e., is the effect of an action or an exogenous event) then we speak of an {\em update}. 
The updated belief base describes the world after its evolution.
The update therefore corresponds to a {\em progression}. \index{progression}

The difference can be illustrated by the following example \cite{Morreau92}. 
Suppose there is a basket containing either an apple or a banana.
If we learn that in fact it does not contain bananas then our beliefs have to be revised and 
we deduce that the basket contains an apple.
However, if we learn that the world has evolved in a way such that there is no banana in the basket any more
(e.g.\ because somebody has performed the action of taking the banana out of the basket if it was there)
then we have to update our beliefs, i.e., that now the basket is either empty or contains an apple.

Just as for revision, there does not exist a unique update operator 
that would suit all applications. 
It is therefore interesting to define criteria which determine which of these operators are `rational', these criteria can be written under the form of rationality postulates\index{rationality postulates}.
Paralleling Alchourr\'on, G\"ardenfors and Makinson's postulates characterizing `rational' revision operators (the so-called AGM postulates) \shortcite{AGM85}, 
Winslett \shortcite{Winslett90} was the first to define postulates for update operators.
These postulates inspired Katsuno and Mendelzon \shortcite{KaMe91} who defined a new set of postulates.
Similarly to the AGM postulates (see Chapter 14 of this volume), they are related to 
the existence of a set of preorder relations \index{order} \index{relation!preference}
about the states of the system, where with each state there is associated a preorder.
In \cite{KaMe91}, the authors implement an idea that had been put forward by Winslett: 
in order to update a belief base one may update each of the models of the base independently.
Katsuno and Mendelzon's contribution is the idea that each model has to be updated towards 
the `closest' models (in the sense of the preorder associated to the original model).
According to Katsuno and Mendelzon, an update operator\index{update} is a function $\diamond$
which given 
a formula ${\cal K}$ representing beliefs about the world and 
a formula $\phi$ representing the information about the evolution of the world,
returns a new formula ${\cal K} \diamond \phi$. The postulates
\index{postulate!of rationality}
they propose in order to characterize the `rational' operators $\diamond$ are the following:\\

\noindent \textbf{U1} ${\cal K} \diamond \varphi$ implies $\varphi$.\\
\textbf{U2} If ${\cal K}$ implies $\varphi$ then $({\cal K} \diamond \varphi)$ is equivalent to ${\cal K}$.\\
\textbf{U3} If ${\cal K}$ and $\varphi$ are satisfiable then ${\cal K} \diamond \varphi$ is 
satisfiable.\\
\textbf{U4} If ${\cal K}_1$ is equivalent to ${\cal K}_2$ and $\varphi_1$ is equivalent to $\varphi_2$ then
${\cal K}_1 \diamond \varphi_1$ is equivalent to ${\cal K}_2 \diamond \varphi_2$.\\
\textbf{U5} $({\cal K} \diamond \varphi)\wedge \psi$ implies ${\cal K} \diamond (\varphi\wedge \psi)$.\\
\textbf{U6} If ${\cal K} \diamond \varphi_1$ implies $\varphi_2$ and ${\cal K} \diamond \varphi_2$ implies $\varphi_1$ then
${\cal K} \diamond \varphi_1$ is equivalent to ${\cal K} \diamond \varphi_2$.\\
\textbf{U7} If ${\cal K}$ is complete then $({\cal K} \diamond  \varphi_1)\wedge({\cal K} \diamond \varphi_2)$ implies
${\cal K} \diamond (\varphi_1\vee\varphi_2)$.\\
\textbf{U8} $({\cal K}_1\vee {\cal K}_2) \diamond \varphi$ is equivalent to $({\cal K}_1 \diamond \varphi)\vee
({\cal K}_2 \diamond \varphi)$.\\
\textbf{U9} If ${\cal K}$ is complete and $({\cal K} \diamond  \varphi_1)\wedge \varphi_2$ is satisfiable then ${\cal K} \diamond (\varphi_1 \wedge \varphi_2)$ implies
$({\cal K} \diamond \varphi_1) \wedge \varphi_2$.
\medskip

U1 stipulates that $\phi$ is a piece of information describing the world after its evolution
(this is one of Winslett's postulates).
U2 says that if $\phi$ was already true in all states of the system before the update 
then the system does not evolve. 
This is the postulate requiring that inertia has always to be preferred to spontaneous evolution.
It is however not always desirable because it forbids the existence of transitory states,
i.e., states within which the system may not stay because it immediately evolves towards other states.
%
U3 expresses that a consistent representation of the system and of its evolution 
can always be updated in a consistent way (which was also one of Winslett's postulates). 
However, systems may exist where there is no transition between two states:
for example, when $\varphi=dead$ and $\alpha=alive$ then one may wish the update to fail.
So this postulate is not always desirable. 
U8 (also one of Winslett's postulates) means that the update is defined as a progression operator. 
U9 is a restriction of the converse of U5.
We refer to \cite{DDP95,HeRi-AIJ99} for a more detailed critique of these postulates.

The following representation theorem relates these postulates\index{postulate!of rationality} 
to the existence of a set of preorder relations\index{relation!of preference}\index{preference relation}
between states of the world:
\begin{theorem}[Katsuno, Mendelzon]\label{volume-1-chapitre12-theoKatsuno}~
$\diamond$ satisfies U1, U2, U3, U4, U5, U8, U9\footnote{
If U6 and U7 are used instead of U9 then the theorem gives us a faithful preorder that is only partial.
} 
if and only if for every $\omega \in \Omega$ there is a total preorder 
$\leq_{\omega}$ 
such that \\
\begin{tabular}{l}
(1) $\forall \omega'\in \Omega, \quad \omega <_{ \omega} \omega'$
($\leq_{\omega}$ is \og{}faithful\fg{}); \\
(2) $Mod({\cal K} \diamond \varphi)=\bigcup_{ \omega \models {\cal K}} \{ \omega' \models \varphi$ such that
$\forall \omega''\models \varphi, \omega'\leq _{\omega} \omega'' \}$.
\end{tabular}
\end{theorem}
Item (1) means that 
for each model $\omega$ of ${\cal K}$, the models of the update of $\omega$ by $\varphi$ are the models of $\varphi$ that are closest to $\omega$ w.r.t.\ $\leq_{\omega}$, and 
(2) means that the set of models of the update of ${\cal K}$ by 
$\varphi$ is the union of the sets of models resulting from the update of each model of ${\cal K}$ by $\varphi$ (which follows directly from postulate U8).

Numerous update operators\index{update} were proposed in the literature. 
Thanks to the above theorem they can be defined by associating with each state a faithful total preorder relation between states.\index{relation!of preference}
In practice, such a set of preorders is a way of \emph{minimizing change}.
For example, Winslett defined a relation $\leq_{\omega}^{\mbox{\tiny PMA}}$ between states that she called 
`Possible Models Approach', abbreviated PMA. 
It is based on the function $\mbox{diff}_{\mbox{\tiny PMA}}(\omega_1,\omega_2)$
(the set of variables whose value differs between the two states $\omega_1$ and $\omega_2$): $\omega_1 \leq_{\omega}^{\mbox{\tiny PMA}} \omega_2 \quad \Leftrightarrow_{def} \quad \mbox{diff}_{\mbox{\tiny PMA}}(\omega_1,\omega) \subseteq
\mbox{diff}_{\mbox{\tiny PMA}}(\omega_2,\omega)$. 
This relation is faithful and therefore defines an update operator.
The corresponding update operator can also be characterized in terms of independence
(the logical consequences of ${\cal K}$ that are independent of $\varphi$ persist) \cite{Marquis94}.
In the examples of \cite{Morreau92}, the initial beliefs are 
${\cal K}=(banana \wedge \neg apple) \vee (apple \wedge \neg banana)$, 
so there are two models $\omega_1=\{\neg apple, banana\}$ and 
$\omega_2=\{apple,\neg banana\}$. 
When the agent then learns ($\varphi$) that somebody took the banana if it was there
(update by $\neg banana$), the states of the system representing the information $\varphi$ are $\omega_2$ and $\omega_3=\{\neg apple, \neg banana\}$. 
The updated base ${\cal K}\diamond_{\mbox{\tiny PMA}} \varphi$ can be computed by taking the union,
for all models $\omega$ of ${\cal K}$, of the models of $\varphi$ that are closest to $\omega$.
Here, the model of $\varphi$ that is closest to $\omega_1$ for the relation $\leq_{\omega}^{\mbox{\tiny PMA}}$ is $\omega_3$; the model of $\varphi$ closest to $\omega_2$ is $\omega_2$ itself
(because $\leq_{\omega}^{\mbox{\tiny PMA}}$ is faithful).
So the set of models of ${\cal K} \diamond_{\mbox{\tiny PMA}} A$ is $\{\omega_2,\omega_3\}$.
This means that after the update there is either an apple in the basket or  the basket is empty.\index{relation!of preference}

The PMA relation has been refined by assigning priorities\index{priority} to some fluents  \cite{Winslett88}, 
which allows for handling fluents that do not persist in the same way.
Other update operators\index{update} go for increased expressiveness, e.g.\ the one proposed by Cordier and Siegel \shortcite{CoSi95}
which allows for more or less prioritary transition constraints\index{constraint!temporal}\index{transition}. 
These constraints take the form of pairs of formulas $(\varphi, \psi)$ and are satisfied by a pair of models 
$(\omega, \omega')$ when $\omega$ satisfies $\varphi$ and $\omega'$ satisfies $\psi$.
Then $\omega'$ is considered closer to $\omega$ than $\omega''$ 
if the transition $(\omega,\omega')$ violates less prioritary constraints than the transition $(\omega,\omega'')$. \index{transition}

Updates\index{update} {\em \`a la} Katsuno and Mendelzon (and in particular Winslett's PMA \shortcite{Winslett88} but also Forbus' operator \cite{Forbus89})  
are built on minimization of change. 
However, minimization of change is not always desirable for updates. 
In particular, Herzig and Rifi \shortcite{HeRi-AIJ99} have shown that the approaches building on
minimization of change do not allow updates by disjunctions;
more formally, an update operator satisfying the Katsuno-Mendelzon postulates 
\index{postulate!of rationality} cannot handle disjunctions correctly, 
the culprit being postulate U5.

For that reason, several scholars studied update operators that are not built on minimization.
They in particular studied a family of update operators that is based on the concept of {\em dependence}\index{dependence}.
Such updates of a belief base $\beta$ by a formula $\alpha$ consists in 
first forgetting in $\beta$  ``all information concerning $\alpha$''
(leaving the truth values of the variables that are not concerned by the update unchanged), 
and then adding $\alpha$ to the result. 
It remains to work out what ``all information concerning $\alpha$'' means.
Such a kind of relevance is induced by a dependence relation between formulas:
$\alpha$ concerns $\beta$ if and only if $\beta$ depends on $\alpha$.
This approach is general because the notion of dependence between formulas can vary.

Most of the dependence-based approaches to update\index{update} consider that the dependence relation is expressed first between formulas and propositional variables, and can then be extended to a dependence relation between formulas: $\alpha$ and $\beta$ are dependent if and only if there is at least one variable on which $\alpha$ and $\beta$ are dependent. 
Examples of such update operators can be found in 
\cite{Herzig96}, \cite{Doherty98} et \cite{HeRi-AIJ99}.
This principle allows for remediating several counterintuitive aspects of 
minimization-based approaches and moreover is generally of lower computational complexity. 
A slight drawback is however that it is too little conservative:
too much information of the initial base is forgotten. 
This can be counterbalanced by replacing the dependence relation between formulas and  variables
by a dependence relation between formulas and literals \cite{HerzigLM13}.

As an update by a formula $\alpha$ can be viewed as a progression \index{progression} 
by a particular action whose effect is $\alpha$ (``to make $\alpha$ true'') 
(see a discussion in \cite{Lang07}), it makes sense to situate update w.r.t.\ propositional action languages. 
We start by observing that STRIPS \index{STRIPS} is a particular case of both formalisms,
corresponding to an update by conjunctions of literals. 
Axiom U8---which requires that the update of a set of models is the union of the update of the individual models---is exactly 
the definition of the progression of a belief state by an action.
The two paradigms however differ in the variety of available actions:
on the one hand, update offers the possibility of taking into account disjunctive effects 
(representing a unique but imperfectly known effect) and more generally effects consisting of arbitrary propositional formulas.
On the other hand, action languages\index{action languages}
allow for conditional effects such as
\begin{center}
({\sf if} Heads {\sf then } flip-coin {\sf causes} $\neg $Tails,  {\sf if} $\neg$Heads {\sf then} flip-coin {\sf
  causes} Tails),
 \end{center}
nondeterministic effects
such as 
\begin{center}
Toss-coin {\sf causes} Heads {\sf or} {\sf causes} $\neg$Heads,
\end{center}
concurrent effects such as
\begin{center}
if $G$ and $D$ are actions consisting in lifting the left and the right side of a table and if a glass of water is on the table then 

$G$ {\sf causes} Spilled,
$D$ {\sf causes} Spilled,  
$G$ {\sf concurrently with} $D$ {\sf cause} $\top$
\end{center}
as well as static causal rules allowing for ramifications.\index{causal rule}\index{ramification problem}
Approaches aiming at unifying the potentialities of various approaches are not numerous.  
Some update approaches take ramification into account by resorting to integrity constraints
\index{integrity constraint} \cite{Doherty98}
or allow for nondeterministic updates \cite{BrewkaHertzberg93}, 
or conditional or concurrent updates \cite{HLMP01}.
However, an embedding of Winslett's and Forbus' update operator and of Dalal's revision operator into dynamic logic was recently provided in \cite{Herzig14}.

Update\index{update} corresponding to the  progression of an action, \index{progression}
there exists a generalization (rightly called {\em generalized update})
that enables both revision \index{revision} and event abduction \cite{Boutilier98}. 
Generalized update allows for example for handling the following scenario:
an agent wakes up in the morning and believes that the lawn is dry just as it was when she went to sleep.
She subsequently observes that the lawn is wet, which first of all leads to a revision of her beliefs, then to the abduction of an event (it rained), and finally to an update by the effects of the event (the road is wet, too).
Belief extrapolation \cite{DupinLang11} and other related formalisms such as \cite{BergerLehmannSchlechta99} only handle the abduction\index{abduction} of events. 
Finally, update can be viewed as an ordinal form of Lewis's {\em imaging} operator \cite{DuboisPrade93} as well as the predictive phase of the Kalman filter\index{Kalman filter}
\cite{CossartTessier99,BenferhatDuboisPrade00}.

Goldszmidt and Pearl \shortcite{GoPe92} were also interested by accounting for revision \index{revision} and update \index{update} at the same time. 
They reason about a set of default rules\index{default} and are of the  causal\index{causal rule}\index{causality} kind,
from which they deduce an order on the pairs of states of the world that they are filtering according to the input.
If the last operator is a revision by $\varphi$ then the pairs of states where the final state satisfies 
$\varphi$ see their plausibility increased. 
In the case of an update by $\varphi$, one has to perform a revision by the dummy action $do(\varphi)$.

\bigskip
The contributions of Winslett, and of Katsuno and Mendelzon, are important for two reasons.
First of all, they established a clear distinction between revising and updating a belief base. 
Second, they elaborated a set of postulates \index{postulate!of rationality}
guaranteeing that a rational update is related to the existence of a set of preorders between the possible states of the system.
Katsuno and Mendelzon, and Winslett, implicitly opted for the particular case where the fluents are by default 
inertial (they only change if an action or an event occurs that changes them).
There is a further implicit hypothesis embodied by postulate U3 (Winslett's MB4):
asserting that any update can be performed means that the input is always consistent with the possible evolution of the world.

Recently, belief change (including belief update) within the framework of fragments of propositional logic has gained attention.
A propositional fragment simply is a subset of a propositional language which has some valuable properties (typically, from the computational side)
but is not fully expressive w.r.t. propositional logic (some propositional formulas do not have any equivalent representation in the fragment).
For instance, the Horn CNF fragment is the set of CNF formulas where each clause is Horn, i.e., it contains at most one positive literal. 
It is well-known that the satisfiability of any Horn CNF formula can be decided in linear time but that some propositional formulas (e.g. the clause $a \vee b$)
cannot be turned into equivalent Horn CNF ones. Other fragments which are often considered are the
Krom one (the set of all CNF formulas where each clause is binary) and the affine fragment (the set of all conjunctions of exclusive-or clauses), and each of them 
offers the same tractability property as the Horn one w.r.t. the satisfiability issue and the same limitations as to expressiveness. 
In order to preserve the tractability benefits, when a belief base from a given fragment has to be updated, it is expected
that the updated base belongs to the same fragment. However, update operators satisfying all the Katsuno-Mendelzon postulates
(especially Winslett's PMA and Forbus' operator) do not ensure this property.  This calls for a notion of refinement of an update operator for a given fragment,
which warrants that the result of any update is in the fragment when the initial base is in the fragment as well. Any refined operator is
required to approximate the behavior of the operator considered at start (especially, leading to the same updated base as it when this base 
fits in the fragment), the price to be paid being the loss of some rationality postulates. A constructive approach to such refinements of update 
operators has been introduced in \cite{CreignouKtariPapini15}, and the Katsuno-Mendelzon postulates satisfied by the refined operators
identified as well.

Several tentatives were also made to extend update to more expressive frameworks:
\begin{itemize}
\item {\em ASP}: 
Slota and Leite  [\citeyear{SlLe10}] adapt Katsuno \& Mendelzon's postulates to logic programs \cite{SlLe10}. They also define update operators for hybrid belief bases \cite{SLS11,Slota12}.
Such hybrid bases are made up of an ontology component, expressed in the language of the description logic ALCIO  (ALC with inverse and nominals),
and a rule component, expressed in the language answer-set programming under the stable semantics (see Chapter 4 of Volume 2).
Update operators are studied in particular for the strong equivalence semantics for ASP
as well as for hybrid belief bases \cite{MoRo10}.
\item 
{\em Belief states}:
Lang, Marquis and Williams [\citeyear{LMW01}] define update operators over epistemic states.
In addition to beliefs, such epistemic states, represented by orders on worlds, allow for expressing the relative plausibility of beliefs.
The authors extend the class of dependence-based update operators to epistemic states. 
Baral and Zhang [\citeyear{BaralZhang05}] generalize update so as to distinguish between facts and knowledge, as in the epistemic logic S5. Their process is called {\em knowledge update} and allows one to account for the effects of epistemic actions by  an {\em updating} the epistemic formulas describing the agent's beliefs.  
Such a framework takes the viewpoint of a modeler agent O who reasons about the belief state of another agent ag. 
For example, the update of an S5 model by $K_{ag} \varphi$ means that O updates 
her beliefs about $ag$'s beliefs; the mental state of $ag$ is seen by O as part of the external world, 
and the update by $K_{ag} \varphi$ corresponds to an action whose effect is to make $K_{ag} \varphi$ true (for example, the action of telling $ag$ that $\varphi$ is true).
\item 
{\em Description logics} (see Chapter 6 of this volume):
Liu et al. [\citeyear{LiuLMW11}] update assertions of an  ``ABox'' (that is, the factual component of the belief base).
They highlight an expressiveness problem that arises in that framework:
sometimes the expected result of an update cannot be encoded by an assertion of the basic description logic ALC.
For example, the assertion $$\mathtt{mary:Person}\sqcap \exists
\mathtt{child\_of}.\mathtt{Person}\sqcap \forall \mathtt{child\_of}.(\mathtt{Person}\sqcap
\mathtt{Happy})$$
expresses that every child of Mary is happy.
If one updates the  ABox containing this information by the fact that Peter becomes unhappy, i.e., by the assertion $\mathtt{peter}:\mathtt{Person} \sqcap \lnot \mathtt{Happy}$,
then one has to take two possible cases into account:
the case where Peter is among Mary's children and the case where he is not. 
Intuitively, the principle of minimal change requires that the result of the update is on the one hand
the new piece of information (Peter is unhappy) and on the other hand the fact that every child of Mary either has the property of being happy or has the property of being Peter. 
The latter assertion (i.e., Mary's children are either happy or called Peter)
cannot be expressed in ALC:
it requires an extension by object names. 
The extended logic (called ALCO) allows for writing 
$$\mathtt{mary:Person}\sqcap \exists
\mathtt{child\_of}.\mathtt{Person}\sqcap \forall \mathtt{child\_of}.(\mathtt{Person}\sqcap
(\mathtt{Happy} \sqcup \{\mathtt{pierre}\}).$$
It turns out that almost all description logic have similar expressiveness problems. 
If we allow for object names as concepts as done in ALCO, this 
wipes out the distinction between ABox assertions (which are about particular objects)
and TBox concept inclusions which should not be about particular objects.
This is unsatisfactory because the distinction is one of the very basic ideas of description logics. 
\item 
{\em Action descriptions}: Eiter et al.  \cite{EiterEFS10} define a framework for minimal change of action descriptions that they call ``action description updates''.
\item 
{\em Abstract argumentation}: researchers in that domain (see Chapter 13 of this volume) are also interested by update and more generally change operators. 
An abstract argumentation system is represented by a graph whose vertices are arguments and whose edges are attacks between arguments. 
Some authors \cite{Boella2009-sh,Cayrol2010-sh,Liao2011-sh,BKRT13,coste2013revision}
are interested by the impact of a change (by adding / with\-drawing arguments or attacks) on such systems.
Baumann and Brewka [\citeyear{Baumann2010-sh,Baumann2012}]
introduced the notion of {\it enforcement}, which is very similar to the notion of update
(cf.\ \cite{Bi2013.8,DBCL16})
because the idea is to minimally modify an argumentation system in a way such that it satisfies a given goal
(usually expressed in terms of arguments that should be accepted).
They define a preference relation between argumentation systems, which is similar to preference relations between models in classical update.
\end{itemize}

\section{Conclusion}

%
%

Reasoning about action and change is one of the oldest topics
 in Artificial Intelligence. Since 1995, the topic is the
subject of a biennial workshop {\em International
  Workshop on Nonmonotonic Reasoning, Action and Change (NRAC)} held
in conjunction with the IJCAI ({\it International Joint Conference on
Artificial Intelligence}) conference.

Several periods followed,
during which the researchers were interested in conceiving  
several formal settings for modeling the important tasks related to 
reasoning about action: STRIPS first, then approaches based on minimal change, and
then approaches based on {\em successor state axioms}. To this
variety of formal settings corresponds a variety of languages \index{language!action}
for representing and reasoning about action: propositional logic, situation calculus, dynamic logic, 
graphical models (among others).

Reasoning about action and change
has close connections with other areas of Artificial Intelligence, including 
non-monotonic reasoning, belief change, reasoning under uncertainty, planning (and in particular
Markov decision processes) \index{Markov Decision Process}; it also has links
with control theory \index{automatic control} (more precisely, Kalman filtering \index{Kalman filter} and discrete event systems).

\bibliography{biblio-vol1-chap12}

\begin{thebibliography}{}

\bibitem[Alchourr\'on et~al., 1985]{AGM85}
Alchourr\'on, C., G{\"a}rdenfors, P., and Makinson, D. (1985).
\newblock On the logic of theory change : partial meet contraction and revision
  functions.
\newblock {\em J. of Symbolic Logic}, 50:510--530.

\bibitem[Aucher and Bolander, 2013]{AucherB13}
Aucher, G. and Bolander, T. (2013).
\newblock Undecidability in epistemic planning.
\newblock In Rossi, F., editor, {\em {IJCAI} 2013, Proceedings of the 23rd
  International Joint Conference on Artificial Intelligence, Beijing, China,
  August 3-9, 2013}, pages 27--33. {IJCAI/AAAI}.

\bibitem[Baral and Zhang, 2005]{BaralZhang05}
Baral, C. and Zhang, Y. (2005).
\newblock Knowledge updates: Semantics and complexity issues.
\newblock {\em Artif. Intell.}, 164(1-2):209--243.

\bibitem[Batusov and Soutchanski, 2018]{BaSo18}
Batusov, V. and Soutchanski, M. (2018).
\newblock Situation calculus semantics for actual causality.
\newblock In {\em Proceedings of the Thirty-Second {AAAI} Conference on
  Artificial Intelligence, New Orleans, Louisiana, USA, February 2-7, 2018}.

\bibitem[Baumann, 2012]{Baumann2012}
Baumann, R. (2012).
\newblock {What Does it Take to Enforce an Argument? Minimal Change in Abstract
  Argumentation}.
\newblock In {\em Proc. European Conf. on Artificial Inteligence (ECAI'12)},
  pages 127--132.

\bibitem[Baumann and Brewka, 2010]{Baumann2010-sh}
Baumann, R. and Brewka, G. (2010).
\newblock {Expanding Argumentation Frameworks: Enforcing and Monotonicity
  Results}.
\newblock In {\em Proc. Int. Conf. on Computational Models of Arugement
  (COMMA'10)}, pages 75--86.

\bibitem[Benferhat et~al., 2000]{BenferhatDuboisPrade00}
Benferhat, S., Dubois, D., and Prade, H. (2000).
\newblock Kalman-like filtering in a possibilistic setting.
\newblock In {\em Proc. European Conf. on Artificial Intelligence (ECAI'00)},
  pages 8--12.

\bibitem[Berger et~al., 1999]{BergerLehmannSchlechta99}
Berger, S., Lehmann, D.~J., and Schlechta, K. (1999).
\newblock Preferred history semantics for iterated updates.
\newblock {\em J. of Logic and Computation}, 9(6):817--833.

\bibitem[Bisquert et~al., 2013]{Bi2013.8}
Bisquert, P., Cayrol, C., Dupin~de Saint~Cyr, F., and Lagasquie-Schiex, M.-C.
  (2013).
\newblock {Enforcement in Argumentation is a kind of Update}.
\newblock In Liu, W., Subrahmanian, V., and Wijsen, J., editors, {\em
  {International Conference on Scalable Uncertainty Management (SUM),
  Washington DC, USA, 16/09/2013-18/09/2013}}.

\bibitem[Boella et~al., 2009]{Boella2009-sh}
Boella, G., Kaci, S., and van~der Torre, L. (2009).
\newblock Dynamics in argumentation with single extensions: {Attack} refinement
  and the grounded extension.
\newblock In {\em Proc. Int. Conf. on Autonomous Agents and Multiagent Systems
  (AAMAS'09)}, pages 1213--1214.

\bibitem[Bolander et~al., 2015]{BolanderJS15}
Bolander, T., Jensen, M.~H., and Schwarzentruber, F. (2015).
\newblock Complexity results in epistemic planning.
\newblock In Yang, Q. and Wooldridge, M., editors, {\em Proceedings of the
  Twenty-Fourth International Joint Conference on Artificial Intelligence,
  {IJCAI} 2015, Buenos Aires, Argentina, July 25-31, 2015}, pages 2791--2797.
  {AAAI} Press.

\bibitem[Booth et~al., 2013]{BKRT13}
Booth, R., Kaci, S., Rienstra, T., and van~der Torre, L. (2013).
\newblock A logical theory about dynamics in abstract argumentation.
\newblock In Liu, W., Subrahmanian, V., and Wijsen, J., editors, {\em
  {International Conference on Scalable Uncertainty Management (SUM),
  Washington DC, USA, 16/09/2013-18/09/2013}}.

\bibitem[Boutilier, 1998]{Boutilier98}
Boutilier, C. (1998).
\newblock A unified model of qualitative belief change: A dynamical systems
  perspective.
\newblock {\em Artificial Intelligence}, 98(1-2):281--316.

\bibitem[Brewka and Hertzberg, 1993]{BrewkaHertzberg93}
Brewka, G. and Hertzberg, J. (1993).
\newblock How to do things with worlds: On formalizing actions and plans.
\newblock {\em J. of Logic and Computation}, 3(5):517--532.

\bibitem[Castilho et~al., 1999]{CGH99}
Castilho, M., Gasquet, O., and Herzig, A. (1999).
\newblock Formalizing action and change in modal logic {I}: The frame problem.
\newblock {\em J. of Logic and Computation}, 9(5):701--735.

\bibitem[Cayrol et~al., 2010]{Cayrol2010-sh}
Cayrol, C., {Dupin de Saint-Cyr}, F., and Lagasquie-Schiex, M.-C. (2010).
\newblock Change in abstract argumentation frameworks: Adding an argument.
\newblock {\em J. of Artificial Intelligence Research}, 38:49--84.

\bibitem[Cooper et~al., 2016]{CooperHMMR16}
Cooper, M.~C., Herzig, A., Maffre, F., Maris, F., and R{\'{e}}gnier, P. (2016).
\newblock A simple account of multi-agent epistemic planning.
\newblock In Kaminka, G.~A., Fox, M., Bouquet, P., H{\"{u}}llermeier, E.,
  Dignum, V., Dignum, F., and van Harmelen, F., editors, {\em {ECAI} 2016 -
  22nd European Conference on Artificial Intelligence, 29 August-2 September
  2016, The Hague, The Netherlands - Including Prestigious Applications of
  Artificial Intelligence {(PAIS} 2016)}, volume 285 of {\em Frontiers in
  Artificial Intelligence and Applications}, pages 193--201. {IOS} Press.

\bibitem[Cordier and Siegel, 1995]{CoSi95}
Cordier, M.-O. and Siegel, P. (1995).
\newblock Prioritized transitions for updates.
\newblock In {\em Proc. European Conf. on Symbolic and Qualitative Aspects of
  Reasoning under Uncertainty (ECSQARU'95)}, pages 142--150.

\bibitem[Cossart and Tessier, 1999]{CossartTessier99}
Cossart, C. and Tessier, C. (1999).
\newblock Filtering vs. revision and update: Let us debate!
\newblock In {\em Proc. European Conf. on Symbolic and Qualitative Aspects of
  Reasoning under Uncertainty (ECSQARU'99)}, pages 116--127.

\bibitem[Coste-Marquis et~al., 2013]{coste2013revision}
Coste-Marquis, S., Konieczny, S., Mailly, J.-G., and Marquis, P. (2013).
\newblock On the revision of argumentation systems: Minimal change of arguments
  status.
\newblock {\em Proc. TAFA}.

\bibitem[Creignou et~al., 2015]{CreignouKtariPapini15}
Creignou, N., Ktari, R., and Papini, O. (2015).
\newblock Belief update within propositional fragments.
\newblock In {\em Proc. European Conference on Symbolic and Quantitative
  Approaches to Reasoning with Uncertainty (ECSQARU'15)}, pages 165--174.

\bibitem[Dannenhauer et~al., 2016]{DMC16}
Dannenhauer, D., Munoz-Avila, H., and Cox, M.~T. (2016).
\newblock Informed expectations to guide gda agents in partially observable
  environments.
\newblock In {\em IJCAI}, pages 2493--2499.

\bibitem[Dannenhauer and Cox, 2017]{DaCo17}
Dannenhauer, Z.~A. and Cox, M.~T. (2017).
\newblock Rationale-based visual planning monitors for cognitive systems.
\newblock In {\em Proceedings of the Thirtieth International Florida Artificial
  Intelligence Research Society Conference}, pages 182--185. AAAI Publications.

\bibitem[Dean and Kanazawa, 1989]{DeKa89}
Dean, T. and Kanazawa, K. (1989).
\newblock A model for reasoning about persistence and causation.
\newblock {\em Computational Intelligence}, 5(2):142--150.

\bibitem[Delgrande and Levesque, 2012]{DeLe12}
Delgrande, J.~P. and Levesque, H.~J. (2012).
\newblock Belief revision with sensing and fallible actions.
\newblock In {\em KR}.

\bibitem[Doherty et~al., 1998]{Doherty98}
Doherty, P., Lukaszewicz, W., and Madalinska-Bugaj, E. (1998).
\newblock The {PMA} and relativizing minimal change for action update.
\newblock In {\em Proc. Int. Conf. on Principles of Knowledge Representation
  and Reasoning (KR'98)}, pages 258--269.

\bibitem[Dubois et~al., 1995]{DDP95}
Dubois, D., {Dupin de Saint-Cyr}, F., and Prade, H. (1995).
\newblock Update postulates without inertia.
\newblock In {\em Proc. European Conf. on Symbolic and Qualitative Aspects of
  Reasoning under Uncertainty (ECSQARU'95)}, pages 162--170.

\bibitem[Dubois and Prade, 1993]{DuboisPrade93}
Dubois, D. and Prade, H. (1993).
\newblock Belief revision and updates in numerical formalisms: An overview,
  with new results for the possibilistic framework.
\newblock In {\em Proc. Int. Joint Conf. on Artificial Intelligence
  (IJCAI'93)}, pages 620--625.

\bibitem[Dupin~de Saint~Cyr, 2008]{DupindeSaintCyr08}
Dupin~de Saint~Cyr, F. (2008).
\newblock {Scenario Update Applied to Causal Reasoning}.
\newblock In {\em {International Conference on Principles of Knowledge
  Representation and Reasoning (KR)}}, pages 188--197,
  http://www.aaai.org/Press/press.php. AAAI Press.

\bibitem[{Dupin de Saint Cyr} et~al., 2016]{DBCL16}
{Dupin de Saint Cyr}, F., Bisquert, P., Cayrol, C., and Lagasquie-Schiex, M.-C.
  (2016).
\newblock {Argumentation update in YALLA (Yet Another Logic Language for
  Argumentation)}.
\newblock {\em International Journal of Approximate Reasoning}, 75:57--92.

\bibitem[{Dupin de Saint-Cyr} and Lang, 2011]{DupinLang11}
{Dupin de Saint-Cyr}, F. and Lang, J. (2011).
\newblock Belief extrapolation (or how to reason about observations and
  unpredicted change).
\newblock {\em Artificial Intelligence}, 175(2):760--790.

\bibitem[Eiter et~al., 2010]{EiterEFS10}
Eiter, T., Erdem, E., Fink, M., and Senko, J. (2010).
\newblock Updating action domain descriptions.
\newblock {\em Artif. Intell.}, 174(15):1172--1221.

\bibitem[Fikes and Nilsson, 1971]{FiNi71}
Fikes, R. and Nilsson, N. (1971).
\newblock Strips : A new approach to the application of theorem proving to
  problem solving.
\newblock {\em Artificial Intelligence}, 2:189--208.

\bibitem[Finger, 1987]{Finger87}
Finger, J. (1987).
\newblock {\em Exploiting constraints in design synthesis}.
\newblock PhD thesis, Stanford University, Stanford, CA.

\bibitem[Forbus, 1989]{Forbus89}
Forbus, K. (1989).
\newblock Introducing actions into qualitative simulation.
\newblock In {\em Proc. Int. Joint Conf. on Artificial Intelligence
  (IJCAI'89)}, pages 1273--1278.

\bibitem[Friedman and Halpern, 1994]{FrHa94}
Friedman, N. and Halpern, J. (1994).
\newblock A knowledge based framework for belief change part {II} : revision
  and update.
\newblock In {\em Proc. Int. Conf. on Principles of Knowledge Representation
  and Reasoning (KR'94)}, pages 190--201.

\bibitem[Friedman and Halpern, 1999]{FriedmanHalpern99}
Friedman, N. and Halpern, J.~Y. (1999).
\newblock Modeling belief in dynamic systems, part {II}: Revision and update.
\newblock {\em J. of Artificial Intelligence Research}, 10:117--167.

\bibitem[Geffner, 1990]{Geffner90}
Geffner, H. (1990).
\newblock Causal theories for nonmonotonic reasoning.
\newblock In {\em Proc. National Conf. on Artificial Intelligence (AAAI'90)},
  pages 524--530.

\bibitem[Gelfond and Lifschitz, 1993]{GeLi93}
Gelfond, M. and Lifschitz, V. (1993).
\newblock Representing action and change by logic programs.
\newblock {\em J. of Logic Programming}, 17:301--321.

\bibitem[Ghallab et~al., 1998]{PDDL98}
Ghallab, M., Howe, A., Knoblock, C., McDermott, D., Ram, A., Veloso, M., Weld,
  C., and Wilkins, D. (1998).
\newblock The planning domain definition language.
\newblock Technical report, AIPS-98 Planning Competition.

\bibitem[Giordano et~al., 1998]{GMS98}
Giordano, L., Martelli, A., and Schwind, C. (1998).
\newblock Dealing with concurrent actions in modal action logics.
\newblock In {\em Proc. European Conf. on Artificial Intelligence (ECAI'98)},
  pages 537--541.

\bibitem[Giunchiglia et~al., 2004]{GLLMT04}
Giunchiglia, E., Lee, J., Lifschitz, V., McCain, N., and Turner, H. (2004).
\newblock Nonmonotonic causal theories.
\newblock {\em Artificial Intelligence}, 153:49--104.

\bibitem[Giunchiglia and Lifschitz, 1998]{giu98}
Giunchiglia, E. and Lifschitz, V. (1998).
\newblock An action language based on causal explanation: Preliminary report.
\newblock In {\em Proc. National Conf. on Artificial Intelligence (AAAI'98)},
  pages 623--630.

\bibitem[Goldszmidt and Pearl, 1992]{GoPe92}
Goldszmidt, M. and Pearl, J. (1992).
\newblock Rank-based systems: A simple approach to belief revision, belief
  update, and reasoning about evidence and actions.
\newblock In {\em Proc. Int. Conf. on Principles of Knowledge Representation
  and Reasoning (KR'92)}, pages 661--672.

\bibitem[Hanks and McDermott, 1986]{HaMD86}
Hanks, S. and McDermott, D. (1986).
\newblock Default reasoning, nonmonotonic logics, and the frame problem.
\newblock In {\em Proc. National Conf. on Artificial Intelligence (AAAI'86)},
  pages 328--333.

\bibitem[Hanks and McDermott, 1994]{HaMD94}
Hanks, S. and McDermott, D. (1994).
\newblock Modelling and uncertain world i : symbolic and probabilistic
  reasoning about change.
\newblock {\em Artificial Intelligence}, 66:1--55.

\bibitem[Heni et~al., 2007]{HBABA07}
Heni, A., {Ben Amor}, N., Benferhat, S., and Alimi, A. (2007).
\newblock Dynamic possibilistic networks: Representation and exact inference.
\newblock In {\em Proc. IEEE Int. Conf. on Computational Intelligence for
  Measurement Systems and Applications (CIMSA’07)}, pages 1--8.

\bibitem[Herzig, 1996]{Herzig96}
Herzig, A. (1996).
\newblock The {PMA} revisited.
\newblock In {\em Proc. Int. Conf. on Principles of Knowledge Representation
  and Reasoning (KR'96)}, pages 40--50.

\bibitem[Herzig, 2014]{Herzig14}
Herzig, A. (2014).
\newblock Belief change operations: {A} short history of nearly everything,
  told in dynamic logic of propositional assignments.
\newblock In Baral, C., Giacomo, G.~D., and Eiter, T., editors, {\em Principles
  of Knowledge Representation and Reasoning: Proceedings of the Fourteenth
  International Conference, {KR} 2014, Vienna, Austria, July 20-24, 2014}.
  {AAAI} Press.

\bibitem[Herzig, 2015]{Herzig15}
Herzig, A. (2015).
\newblock Logics of knowledge and action: critical analysis and challenges.
\newblock {\em Autonomous Agents and Multi-Agent Systems}, 29(5):719--753.

\bibitem[Herzig et~al., 2003]{HerzigLangMarquis03}
Herzig, A., Lang, J., and Marquis, P. (2003).
\newblock Action representation and partially observable planning using
  epistemic logic.
\newblock In {\em Proc. Int. Joint Conf. on Artificial Intelligence
  (IJCAI'03)}, pages 1067--1072.

\bibitem[Herzig et~al., 2013]{HerzigLM13}
Herzig, A., Lang, J., and Marquis, P. (2013).
\newblock Propositional update operators based on formula/literal dependence.
\newblock {\em {ACM} Trans. Comput. Log.}, 14(3):24:1--24:31.

\bibitem[Herzig et~al., 2001]{HLMP01}
Herzig, A., Lang, J., Marquis, P., and Polacsek, T. (2001).
\newblock Updates, actions, and planning.
\newblock In {\em Proc. Int. Joint Conf. on Artificial Intelligence
  (IJCAI'01)}, pages 119--124.

\bibitem[Herzig and Rifi, 1999]{HeRi-AIJ99}
Herzig, A. and Rifi, O. (1999).
\newblock Propositional belief base update and minimal change.
\newblock {\em Artificial Intelligence}, 115(1):107--138.

\bibitem[Herzig and Varzinczak, 2007]{HerzigVarzinczak-Aij07}
Herzig, A. and Varzinczak, I.~J. (2007).
\newblock Metatheory of actions: beyond consistency.
\newblock {\em Artificial Intelligence}, 171:951--984.

\bibitem[Hunter and Delgrande, 2015]{HuDe15}
Hunter, A. and Delgrande, J. (2015).
\newblock Belief change with uncertain action histories.
\newblock {\em Journal of Artificial Intelligence Research}, 53:779--824.

\bibitem[Katsuno and Mendelzon, 1991]{KaMe91}
Katsuno, H. and Mendelzon, A. (1991).
\newblock On the difference between updating a knowledge base and revising it.
\newblock In {\em Proc. Int. Conf. on Principles of Knowledge Representation
  and Reasoning (KR'91)}, pages 387--394.

\bibitem[Keller and Winslett, 1985]{KeWi85}
Keller, A. and Winslett, M. (1985).
\newblock On the use of an extended relational model to handle changing
  incomplete information.
\newblock In {\em IEEE Trans. on Software Engineering}, volume SE-11:7, pages
  620--633.

\bibitem[Lang, 2007]{Lang07}
Lang, J. (2007).
\newblock {Belief Update Revisited}.
\newblock In {\em Proc. Int. Joint Conf. on Artificial Intelligence
  (IJCAI'07)}, pages 2517--2522.

\bibitem[Lang et~al., 2003]{LangLM03}
Lang, J., Liberatore, P., and Marquis, P. (2003).
\newblock Propositional independence: Formula-variable independence and
  forgetting.
\newblock {\em J. of Artificial Intelligence Research}, 18:391--443.

\bibitem[Lang et~al., 2001]{LMW01}
Lang, J., Marquis, P., and Williams, M.-A. (2001).
\newblock Updating epistemic states.
\newblock In {\em Proc. Australian Joint Conf. on Artificial Intelligence
  (AI'01)}, pages 297--308.

\bibitem[Li et~al., 2017]{LiYW17}
Li, Y., Yu, Q., and Wang, Y. (2017).
\newblock More for free: a dynamic epistemic framework for conformant planning
  over transition systems.
\newblock {\em J. Log. Comput.}, 27(8):2383--2410.

\bibitem[Liao et~al., 2011]{Liao2011-sh}
Liao, B., Jin, L., and Koons, R.~C. (2011).
\newblock Dynamics of argumentation systems: A division-based method.
\newblock {\em Artificial Intelligence}, 175(11):1790 -- 1814.

\bibitem[Lifschitz, 1990]{Lifschitz90}
Lifschitz, V. (1990).
\newblock Frames in the space of situations.
\newblock {\em Artificial Intelligence}, 46:365--376.

\bibitem[Lifschitz and Rabinov, 1989]{LiRa89}
Lifschitz, V. and Rabinov, A. (1989).
\newblock Things that change by themselves.
\newblock In {\em Proc. Int. Joint Conf. on Artificial Intelligence
  (IJCAI'89)}, pages 864--867.

\bibitem[Lin, 1995]{Lin95}
Lin, F. (1995).
\newblock Embracing causality in specifying the indirect effects of actions.
\newblock In {\em Proc. Int. Joint Conf. on Artificial Intelligence
  (IJCAI'95)}, pages 1985--1991.

\bibitem[Liu et~al., 2011]{LiuLMW11}
Liu, H., Lutz, C., Milicic, M., and Wolter, F. (2011).
\newblock Foundations of instance level updates in expressive description
  logics.
\newblock {\em Artificial Intelligence}, 175(18):2170--2197.

\bibitem[Marquis, 1994]{Marquis94}
Marquis, P. (1994).
\newblock Possible models approach via independency.
\newblock In {\em Proc. European Conf. on Artificial Intelligence (ECAI'94)},
  pages 336--340.

\bibitem[McCarthy, 1977]{McCarthy77}
McCarthy, J. (1977).
\newblock Epistemological problems of artificial intelligence.
\newblock In {\em Proc. Int. Joint Conf. on Artificial Intelligence
  (IJCAI'77)}, pages 1038--1044.

\bibitem[McCarthy, 1986]{McCarthy86}
McCarthy, J. (1986).
\newblock Applications of circumscription to formalizing common-sense
  knowledge.
\newblock {\em Artificial Intelligence}, 28(1):1038--1044.

\bibitem[McCarthy and Hayes, 1969]{McHa69}
McCarthy, J. and Hayes, P. (1969).
\newblock Some philosophical problems from the standpoint of artificial
  intelligence.
\newblock In {\em Machine Intelligence}, volume~4, pages 463--502.

\bibitem[Moinard, 2000]{Moinard00}
Moinard, Y. (2000).
\newblock Note about cardinality-based circumscription.
\newblock {\em Artificial Intelligence}, 119(1-2):259--273.

\bibitem[Molineaux and Aha, 2014]{MoAh14}
Molineaux, M. and Aha, D.~W. (2014).
\newblock Learning unknown event models.
\newblock In {\em AAAI}, pages 395--401.

\bibitem[Morreau, 1992]{Morreau92}
Morreau, M. (1992).
\newblock Planning from first principles.
\newblock In {\em Belief revision}, pages 204--219. Cambridge University Press.

\bibitem[Motik and Rosati, 2010]{MoRo10}
Motik, B. and Rosati, R. (2010).
\newblock Reconciling description logics and rules.
\newblock {\em J. of the Association for Computing Machinery}, 57(5).

\bibitem[Pearl, 1988]{Pearl88}
Pearl, J. (1988).
\newblock Embracing causality in formal reasoning.
\newblock {\em Artificial Intelligence}, 35:259--271.

\bibitem[Pednault, 1989]{Pednault89}
Pednault, E. (1989).
\newblock Adl: Exploring the middle ground between strips and the situation
  calculus.
\newblock In {\em Proc. Int. Conf. on Principles of Knowledge Representation
  and Reasoning (KR'89)}, pages 324--332.

\bibitem[Reiter, 1991]{Reiter91}
Reiter, R. (1991).
\newblock The frame problem in the situation calculus: A simple solution
  (sometimes) and a completeness result for goal regression.
\newblock In {\em Artificial Intelligence and Mathematical Theory of
  Computation: Papers in Honor of John McCarthy}, pages 359--380. Academic
  Press.

\bibitem[Sandewall, 1995]{Sandewall95}
Sandewall, E. (1995).
\newblock {\em Features and Fluents}.
\newblock Oxford University Press.

\bibitem[Scherl and Levesque, 2003]{ScherlLevesque03}
Scherl, R. and Levesque, H.~J. (2003).
\newblock The frame problem and knowledge producing actions.
\newblock {\em Artificial Intelligence}, 144(1-2).

\bibitem[Shoham, 1988]{Shoham88}
Shoham, Y. (1988).
\newblock {\em Reasoning about Change - Time and Causation from the Standpoint
  of Artificial Intelligence}.
\newblock MIT Press.

\bibitem[Slota, 2012]{Slota12}
Slota, M. (2012).
\newblock {\em Updates of Hybrid Knowledge bases}.
\newblock PhD thesis, Universidade Nove de Lisboa.

\bibitem[Slota and Leite, 2010]{SlLe10}
Slota, M. and Leite, J. (2010).
\newblock On semantic update operators for answer-set programs.
\newblock In {\em Proc. European Conf. on Artificial Intelligence}, pages
  957--962.

\bibitem[Slota et~al., 2011]{SLS11}
Slota, M., Leite, J., and Swift, T. (2011).
\newblock Splitting and updating hybrid knowledge bases.
\newblock {\em Theory and Practice of Logic Programming}, 11(4-5):801--819.

\bibitem[Stein and Morgenstern, 1994]{StMo94}
Stein, L. and Morgenstern, L. (1994).
\newblock Motivated action theory: A formal theory of causal reasoning.
\newblock {\em Artificial Intelligence}, 71(1):1--42.

\bibitem[Thielscher, 1995]{Thielscher95}
Thielscher, M. (1995).
\newblock The logic of dynamic systems.
\newblock In {\em Proc. Int. Joint Conf. on Artificial Intelligence
  (IJCAI'95)}, pages 1956--1962.

\bibitem[Thielscher, 1997]{Thielscher97}
Thielscher, M. (1997).
\newblock Ramification and causality.
\newblock {\em Artificial Intelligence}, 89(1–2):317–364.

\bibitem[Turner, 1999]{Turner99}
Turner, H. (1999).
\newblock A logic of universal causation.
\newblock {\em Artificial Intelligence}, 113(1-2):87--123.

\bibitem[{van Ditmarsch} et~al., 2011]{DitmarschHerzigLima-Jlc10}
{van Ditmarsch}, H., Herzig, A., and de~Lima, T. (2011).
\newblock {From Situation Calculus to Dynamic Epistemic Logic}.
\newblock {\em J. of Logic and Computation}, 21(2):179--204.

\bibitem[Winslett, 1988]{Winslett88}
Winslett, M. (1988).
\newblock Reasoning about action using a possible models approach.
\newblock In {\em Proc. National Conf. on Artificial Intelligence (AAAI'88)},
  pages 89--93.

\bibitem[Winslett, 1990]{Winslett90}
Winslett, M. (1990).
\newblock {\em Updating Logical Databases}.
\newblock Cambridge University Press.

\end{thebibliography}
\bibliographystyle{apalike}


\end{document}